\documentclass{article}
\def\GitHub{\url{https://github.com/magic-research/vector_quantization}}

\def\ImageNetOneK{ImageNet-1k}
\def\INOneK{IN-1k}
\def\MSCOCO{MS-COCO}

\def\InceptionVThree{Inception-v3}

\def\CLIP{CLIP}
\def\DINO{DINO}
\def\MAE{MAE}
\def\OpenCLIP{OpenCLIP}

\def\ViT{ViT}
\makeatletter
\newcommand{\@ViT}[1]{\ViT-{#1}}
\newcommand{\@@ViT}[2]{\@ViT{#1}/#2}
\def\ViTB{\@ViT{B}}
\def\ViTL{\@ViT{L}}
\def\ViTH{\@ViT{H}}
\def\ViTG{\@ViT{G}}
\def\ViTBSixteen{\@@ViT{B}{16}}
\def\ViTLFourteen{\@@ViT{L}{14}}
\def\ViTHFourteen{\@@ViT{H}{14}}
\def\ViTGFourteen{\@@ViT{G}{14}}
\makeatother

\def\GPT{GPT}
\makeatletter
\newcommand{\@GPT}[1]{\GPT-{#1}}
\newcommand{\@@GPT}[2]{\@GPT{#1} {#2}}
\def\GPTTwo{\@GPT{2}}
\def\GPTTwoMedium{\@@GPT{2}{Medium}}
\def\GPTTwoXL{\@@GPT{2}{XL}}
\makeatother

\def\AR{AR}
\def\AutoRegressive{AutoRegressive}
\def\IG{IG}
\def\IU{IU}
\def\NAR{NAR}

\def\MIM{MIM}
\def\TTwoI{T2I}

\def\TextToImage{Text-to-Image}

\def\BEiT{BEiT}
\def\BEiTvTwo{\BEiT~v2}
\def\CVQVAE{C\VQVAE}

\def\FSQ{FSQ}
\def\GAN{GAN}
\def\HQVAE{HQ-\VAE}
\def\LFQ{LFQ}
\def\MAGE{MAGE}
\def\MUSE{MUSE}
\def\MaskGIT{MaskGIT}
\def\MoVQ{Mo\VQ}
\def\RQVAE{RQ-\VAE}
\def\SQVAE{SQ-\VAE}
\def\SeQGAN{SeQ-\GAN}
\def\VAE{VAE}
\def\VQ{VQ}
\def\VQGAN{\VQ\GAN}
\def\VQVAE{\VQ-\VAE}
\def\VQVAETwo{\VQVAE-2}
\def\VQWAE{\VQ-\WAE}
\def\ViTVQGAN{\ViT-\VQGAN}
\def\WAE{WAE}

\def\VQKD{\VQ-KD}
\makeatletter
\newcommand{\@VQKD}[1]{\VQKD\textsubscript{#1}}
\def\VQKDCLIP{\@VQKD{\CLIP}}
\def\VQKDDINO{\@VQKD{\DINO}}
\def\VQKDMAE{\@VQKD{\MAE}}
\def\VQKDViT{\@VQKD{\ViT}}
\makeatother

\def\Cluster{Cluster}
\makeatletter
\newcommand{\@Cluster}[1]{\Cluster\textsubscript{#1}}
\def\ClusterCLIP{\@Cluster{\CLIP}}
\def\ClusterDINO{\@Cluster{\DINO}}
\def\ClusterMAE{\@Cluster{\MAE}}
\def\ClusterViT{\@Cluster{\ViT}}
\def\ClusterStar{\@Cluster{$\star$}}
\makeatother

\def\FID{FID}
\def\FIDAR{\FID\textsubscript{\AR}}
\def\FIDNAR{\FID\textsubscript{\NAR}}
\def\FIDTTwoI{\FID\textsubscript{\TTwoI}}
\def\FrechetInceptionDistance{Fr\'{e}chet Inception Distance}
\def\IS{IS}
\def\ISAR{\IS\textsubscript{\AR}}
\def\ISNAR{\IS\textsubscript{\NAR}}
\def\PPL{PPL}
\def\rFID{r\FID}

\def\D{\mathcal{D}}
\def\T{\mathcal{T}}
\def\E{\mathcal{E}}
\def\Q{\mathcal{Q}}
\def\C{\mathbf{C}}
\def\z{\mathbf{z}}

\def\P{\mathcal{P}}
\makeatletter
\newcommand{\@P}[1]{\P_{#1}}
\def\PAR{\@P{\text{\AR}}}
\def\PNAR{\@P{\text{\NAR}}}
\makeatother

\newcommand{\sg}[1]{\text{sg}[#1]}

\newif\ifarxiv
\arxivtrue

\ifarxiv
\usepackage[preprint,nonatbib]{neurips_2024}
\else
\usepackage[final,nonatbib]{neurips_2024}
\fi

\usepackage{amsfonts}
\usepackage{booktabs}
\usepackage{hyperref}
\usepackage{nicefrac}
\usepackage{subcaption}
\usepackage{wrapfig}

\usepackage{utils}

\title{Image Understanding Makes for A Good Tokenizer for Image Generation}

\author{
  Luting Wang\thanks{Work done during an internship at ByteDance.}\quad
  Yang Zhao$^1$\quad
  Zijian Zhang$^1$\quad
  Jiashi Feng$^1$\quad
  Si Liu\thanks{Corresponding authors (\href{mailto:liusi@buaa.edu.cn}{liusi@buaa.edu.cn}, \href{mailto:bingyikang@bytedance.com}{bingyikang@bytedance.com}).}\quad
  Bingyi Kang$^{1\dagger}$\thanks{Project lead.}\\
  $^1$ByteDance
}

\begin{document}

\maketitle

\setcounter{footnote}{0}

\begin{abstract}
  Modern image generation (\IG) models have been shown to capture rich semantics valuable for image understanding (\IU) tasks. 
  However, the potential of \IU{} models to improve \IG{} performance remains uncharted. 
  We address this issue using a token-based \IG{} framework, which relies on effective tokenizers to project images into token sequences. 
  Currently, \textit{pixel reconstruction} (\eg, \VQGAN) dominates the training objective for image tokenizers.
  In contrast, our approach adopts the \textit{feature reconstruction} objective, where tokenizers are trained by distilling knowledge from pretrained \IU{} encoders. 
  Comprehensive comparisons indicate that tokenizers with strong \IU{} capabilities achieve superior \IG{} performance across a variety of metrics, datasets, tasks, and proposal networks. 
  Notably, \VQKDCLIP{} achieves $4.10$ \FID{} on \ImageNetOneK{} (\INOneK).
  Visualization suggests that the superiority of \VQKD{} can be partly attributed to the rich semantics within the \VQKD{} codebook.
  We further introduce a straightforward pipeline to directly transform \IU{} encoders into tokenizers, demonstrating exceptional effectiveness for \IG{} tasks. 
  These discoveries may energize further exploration into image tokenizer research and inspire the community to reassess the relationship between \IU{} and \IG.
  The code is released at \GitHub.
\end{abstract}

\section{Introduction}
\label{sec:introduction}

\begin{figure}[t]
  \centering
  \includegraphics[width=\linewidth]{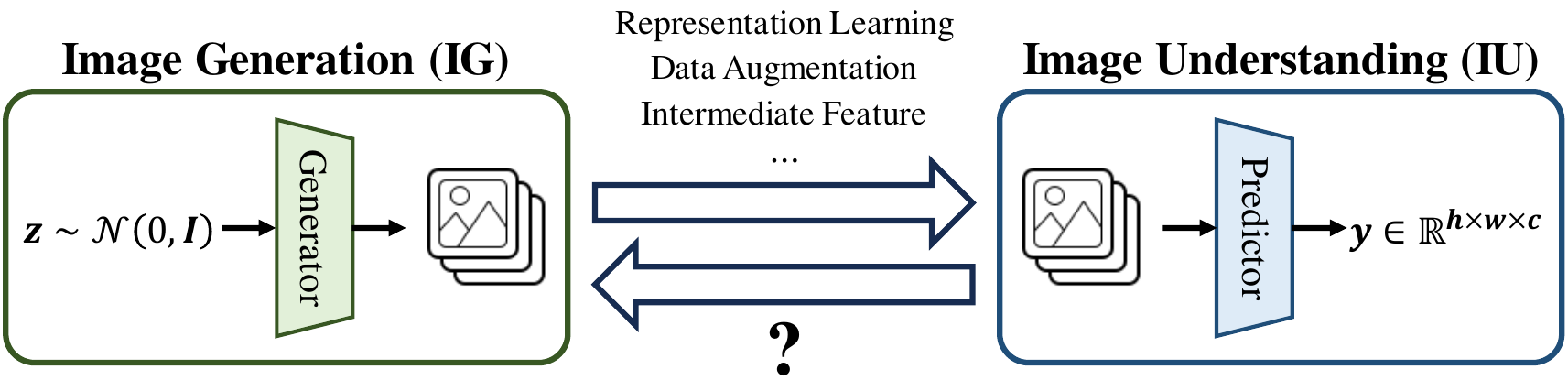}
  \caption{
    Extensive studies have tried to adopt \IG{} models for \IU.
    However, few attempts have been made to use \IU{} models in \IG.
  }
  \label{fig:fig1}
\end{figure}

Image understanding (\IU) and image generation (\IG) have been the core pursuits of computer vision research for a long time. 
Thanks to the progress in generative models~\cite{dpm,ncsn,ddpm,ddim,parti} and network architectures~\cite{transformer,vit}, \IG{} has witnessed remarkable advancements in recent years. These advancements spurred extensive research on leveraging powerful \IG{} models for \IU{} tasks (\Cref{fig:fig1}). 
Studies have shown that \IG{} models can benefit \IU{} tasks in various ways, including data augmentation through synthetic data generation~\cite{diffusionengine,freemask,datasetdm}, improved representation learning~\cite{diff_ae,soda,pdae}, and utilizing intermediate features from \IG{} models for solving perception tasks~\cite{vpd,gd}.
However, the reciprocal question remains largely uncharted: \textit{how might \IU{} models aid \IG{} tasks?}

The primary focus of this paper lies in the \AutoRegressive{} (\AR) \IG{} framework, which is gaining considerable attention for its excellence in generating high-quality images and videos~\cite{parti,cm3leon,videopoet}.
This framework operates in a two-stage process.
The first stage learns a tokenizer to map images into sequences of discrete tokens.
Subsequently, the second stage trains a proposal network to model the token sequences.
As underlined by prior research~\cite{magvitv2,muse}, the quality of the tokenizers significantly influences overall \IG{} performance.
Meanwhile, tokenizers and \IU{} encoders adhere to a similar structure as they both aim to map images into latent representations, either discrete or continuous.
As a result, the token-based \IG{} framework provides an optimal environment for investigating the relationship between \IU{} and \IG{}.
Through comprehensive studies, we demonstrate that existing \IU{} models from representation learning can be useful in generative models, even if they are not specifically designed for the \IG{} task.

Our study involves training three components within the \AR{} framework: tokenizer, decoder, and proposal network. 
Traditionally, \textit{pixel reconstruction} has been the dominant objective for training tokenizers, such as \VQGAN~\cite{vqgan} and \FSQ~\cite{fsq}.  
To the best of our knowledge, we are the first to systematically demonstrate that \textit{feature reconstruction} (\VQKD~\cite{beitv2}) achieves better \IG{} performance. 
\footnote{\VQKD{} was initially proposed for image pretraining.}
This approach distills knowledge from pretrained \IU{} encoders to tokenizers.
Therefore, the training strategy of the \IU{} encoder is crucial for the performance of the tokenizer.
In this regard, we investigate four representative \IU{} encoders: \ViT~\cite{vit}, \CLIP~\cite{clip}, \DINO~\cite{dino}, and \MAE~\cite{mae}.  
Following \VQGAN~\cite{vqgan}, we train decoders to restore pixel values from discrete tokens, and proposal networks (\AR{} or \NAR) that can model the distribution of image tokens.  
The models are then evaluated using various metrics, including codebook usage, \FrechetInceptionDistance{} (\FID)~\cite{fid}, Inception Score (\IS)~\cite{is}, perplexity (\PPL), \etc.

Initially, we compare the above tokenizers on \INOneK{} for class-conditional \IG.
\VQKD{} achieves $4.10$ \FIDAR{}, outperforming \VQGAN{} ($15.78$ \FIDAR) by a large margin.
\FSQ{} experiments confirm that the superiority of \VQKD{} is not solely attributable to the specific quantization operation or high codebook usage.
More generally, \VQKD{} consistently outperforms across different proposal networks, datasets, and tasks.

We analyze \VQKD{} from multiple perspectives.
By visualizing the codebook, we discover that codes from \VQKD{} carry more semantics than \VQGAN, which makes them easier to model and subsequently improve the \IG{} quality.
Building upon this insight, we propose a straightforward pipeline to efficiently transform \IU{} encoders into tokenizers, outperforming \VQKD{} on the \MSCOCO{} dataset.
We also find that tokenizers with weaker \IU{} capabilities require larger proposal networks for effective \AR{} modeling and show less robustness to variations in the training images.
Finally, we conduct qualitative analysis to present the visual results.

In sum, the key insights from our study include the following: 1) This research is the first to demonstrate that \IU{} models can substantially enhance \IG{} through \VQKD; 2) Tokenizers with strong \IU{} capabilities consistently outperform conventional \VQGAN-based methods across various metrics, datasets, tasks, and network architectures; 3) The \VQKD{} codebook encapsulates more semantics than \VQGAN, contributing to the superiority of \VQKD{} in \IG.

We believe these findings can benefit future research on image tokenizers and provoke further discussion on the relationship between \IU{} and \IG.

\section{Related Work}

\paragraph{Image Tokenization.}
Vector Quantization (\VQ)~\cite{vq} is originally developed for data compression.
To circumvent posterior collapse in the \VAE~\cite{vae} framework, \VQVAE~\cite{vq_vae} adopts \VQ{} for image tokenization.
Subsequently, \VQGAN~\cite{vqgan} introduces adversarial and perceptual losses to enhance the quality of the generated images.

Vanilla \VQGAN{} suffers from limitations like low codebook usage, limited semantic representation ability, and the trade-off between modeling efficiency and image quality.
To address these challenges, researchers have focused on improving the codebook.
\ViTVQGAN~\cite{vit_vqgan} adopts a \ViT-based~\cite{vit} autoencoder to create more expressive code vectors.
\SeQGAN~\cite{seq_gan} improves the perceptual loss and decoder to balance between semantic compression and detail preservation.
\SQVAE~\cite{sq_vae} improves \VQVAE{} with stochastic quantization and a trainable posterior categorical distribution.
\VQWAE~\cite{vq_wae} builds upon \SQVAE{} by encouraging the discrete representation to be a uniform distribution via a Wasserstein distance.
\HQVAE~\cite{hq_vae} employs random re-initialization of inactive code vectors.
\CVQVAE~\cite{cvq_vae} selects encoded features as anchors to update dead codes.
\VQKD~\cite{beitv2} adopts knowledge distillation instead of image reconstruction as the objective to train \VQVAE{}.
\LFQ~\cite{magvitv2} and \FSQ~\cite{fsq} adopt bounded scalar quantization techniques from neural compression to harness the potential of extra-large codebooks.

Furthermore, several works explore the potential of multiple codebooks.
\VQVAETwo~\cite{vq_vae_2} extends \VQVAE{} to a multi-scale hierarchical organization.
\RQVAE~\cite{rq_vae} and \MoVQ~\cite{movq} aim to represent each feature as a stack of tokens, where \RQVAE{} adopts an iterative way to factorize features into a series of residuals and \MoVQ{} models features across multiple channels via specialized modulation.

\paragraph{Token-based Image Generation.}
Inspired by the success of \GPT~\cite{gpt,gpt_2,gpt_3}, \VQVAE{} and most of its derivative works~\cite{vqgan, movq} adopt \AR{} transformers to model the token sequence.
This approach leverages techniques from text generation to enhance \IG{} performance.
However, the decoding time of \AR{} models scales linearly with the length of the token sequence.
To accelerate decoding, \MaskGIT~\cite{maskgit} introduces a bidirectional transformer, referred to as the \NAR{} proposal network.

Given the versatility of token-based modeling, both \AR{} and \NAR{} proposal networks can be easily extended to conditional \IG{} scenarios.
For instance, \VQGAN{} uses a class token as the condition in its \AR{} proposal network for class-conditional \IG{}. 
With an \NAR{} proposal network, \MUSE~\cite{muse} adopts text embeddings to predict masked image tokens in \TextToImage{} generation.

\begin{figure}[t]
  \centering
  \includegraphics[width=\linewidth]{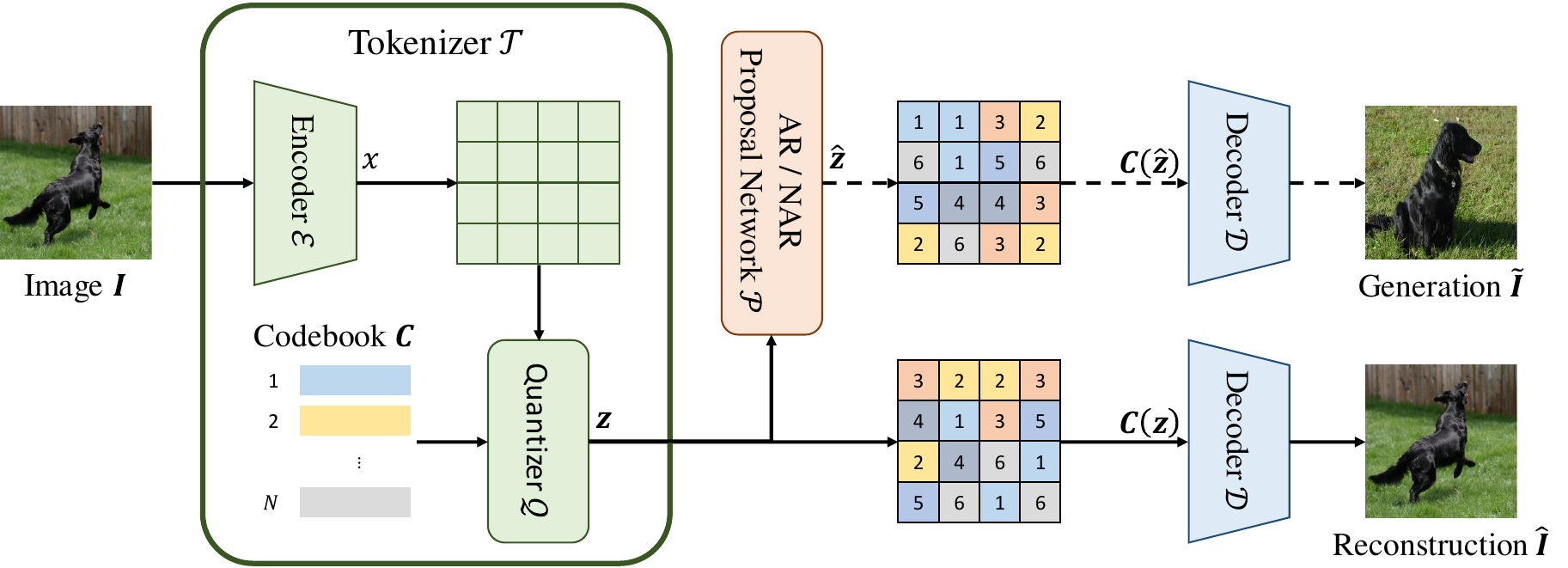}
  \caption{
    The token-based \IG{} framework.
    Solid and dashed lines represent training and inference pipelines, respectively.
    During training, the tokenizer $\T$ tokenizes an image $\II$ into discrete codes $\z$.
    A proposal network $\P$ is trained to model the distribution $p(\z)$, while a decoder $\D$ learns to reconstruct $\II$.
    During inference, we sample codes $\hat{\z}$ from $\P$, which guides $\D$ to perform generation.
  }
  \label{fig:overview}
\end{figure}

\section{Token-Based Image Generation}

\begin{figure}[t]
  \begin{subfigure}[b]{0.3\linewidth}
    \includegraphics[width=\linewidth]{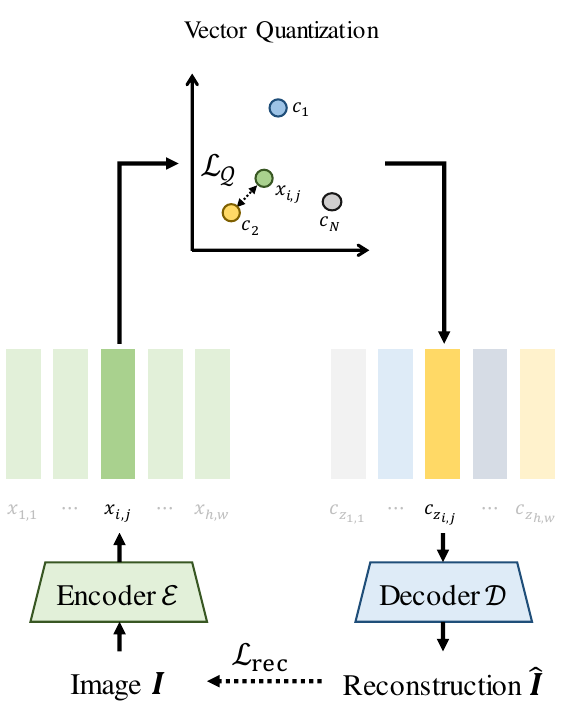}
    \caption{\VQGAN}
    \label{fig:vqgan}
  \end{subfigure}
  \hfill
  \begin{subfigure}[b]{0.3\linewidth}
    \includegraphics[width=\linewidth]{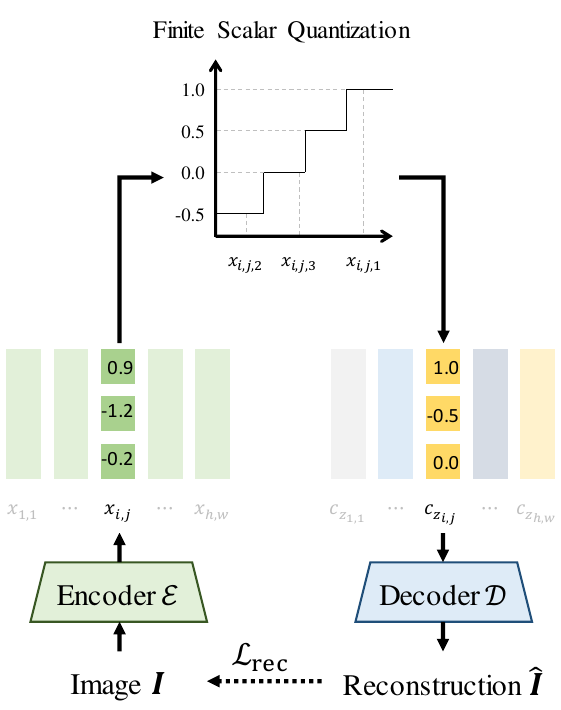}
    \caption{\FSQ}
    \label{fig:fsq}
  \end{subfigure}
  \hfill
  \begin{subfigure}[b]{0.3\linewidth}
    \includegraphics[width=\linewidth]{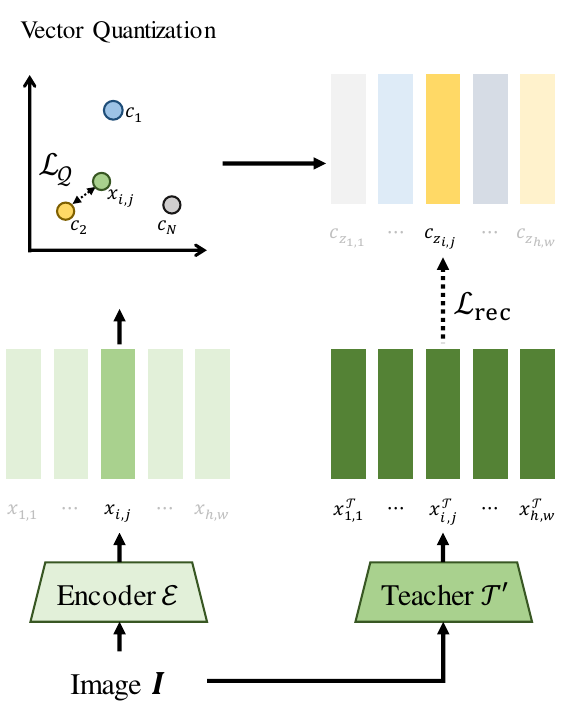}
    \caption{\VQKD}
    \label{fig:vqkd}
  \end{subfigure}
  \caption{
    The architecture and training objective of different image tokenizers.
  }
\end{figure}

We start with the two-stage \IG{} framework in \Cref{sec:two_stage_image_generation}.
Subsequently, \Cref{sec:image_tokenizers} details the architecture and training protocol for the tokenizers under consideration.
\Cref{sec:benchmark} explains the evaluation benchmark.
\Cref{sec:main_observation} further outlines our main observations derived from the \INOneK{} experiments.
Lastly, we validate the observations under different settings in \Cref{sec:further_verification}.

\subsection{Two-Stage Image Generation}
\label{sec:two_stage_image_generation}

We illustrate the two-stage \IG{} framework in \Cref{fig:overview}.
Given an image $\II \in \R^{H \times W \times 3}$, the encoder $\E$ converts this image into a feature map $x \in \R^{h \times w \times d}$, where $(h, w) = (H / f, W / f)$ and $f$ is a downsample factor.
Let a codebook $\C$ be a set of $N$ code vectors $\{c_i\}_{i = 1}^N \in \R^{N \times d}$, where each code vector $c_i \in \R^d$ corresponds to a specific code $i$.
A quantizer $\Q$ then maps $x$ into a sequence of codes $\z = \{z_i\}_{i = 1}^L$, where $L = h \times w$ defines the sequence length and $z_i$ is an integer that falls within the range $[1, N]$.
Let $c_{z_i}$ denote the code vector that corresponds to code $z_i$.
Similarly, $\C(\z) = \{c_{z_i}\}_{i = 1}^L \in \R^{L \times d}$ represents a sequence of code vectors associated with the code sequence $\z$.
The encoder $\E$, quantizer $\Q$, and codebook $\C$ collectively form an image tokenizer $\T$.

The proposal network $\P$ models the distribution over $\z$, where the distribution is denoted as $p(\z)$.
Early proposal networks are implemented as an \AR{} transformer, which sequentially models $p(z_i|z_{1:i-1})$ and formulates $p(\z)$ as $\prod_{i=1}^{h \times w} p(z_i|z_{1:i-1})$.
While the \AR{} transformers can be trained in parallel, it has to sequentially decode $\z$ during inference, which renders it inefficient.
Therefore, \NAR{} proposal networks are being prevalent~\cite{maskgit}, which typically adopt bidirectional transformers to model $\z$.
We denote the two types of proposal networks as $\PAR$ and $\PNAR$, respectively.

Finally, a decoder $\D$ maps the code vectors to the pixel space.
In training, $\D$ takes $\C(\z)$ as input and learns to reconstruct the original image $\II$ as $\hat{\II} \in \R^{H \times W \times 3}$.
In inference, a sequence of codes $\hat{\z}$ is sampled from $p(\z)$, translated to $\C(\hat{\z})$, and then fed into $\D$ to generate an image $\tilde{\II}$.

\subsection{Image Tokenizers}
\label{sec:image_tokenizers}

In this paper, we focus on three types of image tokenizers: \VQGAN~\cite{vqgan}, \FSQ~\cite{fsq}, and \VQKD~\cite{beitv2}.

Let $x_{i, j} \in \R^{d}$ be a vector in the feature map $x$.
As shown in \Cref{fig:vqgan}, to quantize $x_{i, j}$, the \VQGAN{} tokenizer looks up the codebook $\C$ for the closest code vector in terms of Euclidean distance:
\begin{equation}
  z_{i, j} = \arg\min_{z} \lVert x_{i, j} - c_z \rVert_2.
\end{equation}
Since the quantization process is non-differentiable, \VQGAN{} adopts the Straight-Through Estimator (STE)~\cite{ste} to optimize the encoder $\E$, which copies gradients from $\C(\z)$ to $x$.
As a result, the codebook $\C$ receives no gradient.
To optimize $\C$, \VQGAN{} introduces a quantization loss $\LL_{\Q}$:
\begin{equation}
  \LL_{\Q} = \lVert \sg{x} - \C(\z) \rVert_2^2 + \beta \lVert x - \sg{\C(\z)} \rVert_2^2,
\end{equation}
where $\sg{\cdot}$ denotes the stop-gradient operation and $\beta$ is the loss weight.
The first term is the codebook loss, which optimizes the codebook.
The second term is the commitment loss to make sure the encoder $\E$ commits to a code vector~\cite{vq_vae}.
Therefore, the overall loss for \VQGAN{} is defined as:
\begin{equation}
  \LL = \LL_{\Q} + \LL_\text{rec}(\II, \hat{\II}),
\end{equation}
where $\LL_\text{rec}$ is the reconstruction loss between image $\II$ and reconstruction $\hat{\II}$, which includes $\ell_1$ loss, perceptual loss, and adversarial loss.

Based on \VQGAN, \FSQ{} introduces a simpler image tokenizer, without the need for codebook lookup and quantization loss.
As shown in \Cref{fig:fsq}, \FSQ{} adopts finite scalar quantization to quantize each channel of $x_{i, j}$ into a finite set of scalars.
Since the quantization process involves no trainable parameter, \FSQ{} can be trained with solely the reconstruction loss $\LL_\text{rec}(\II, \hat{\II})$.

Unlike \VQGAN{} and \FSQ, which are designed for \IG, \VQKD{} was originally presented in \BEiTvTwo{} to provide supervision for \IU{} models.
As shown in \Cref{fig:vqkd}, \VQKD{} is trained to reconstruct the feature map $x^\T$ encoded by a pretrained teacher $\T^{'}$.
Formally, the reconstruction loss is defined as:
\begin{equation}
  \LL_\text{rec} = -\cos\left(\D(\C(\z)), x^\T\right),
\end{equation}
where $\cos(\cdot, \cdot)$ represents cosine similarity.

In this study, we examine \VQKD{} using four types of pretrained teachers, including fully-supervised, text-supervised, contrastive, and Masked Image Modeling (\MIM).
We use \VQKDCLIP{} and \VQKDDINO{} to represent \VQKD{} tokenizers trained with \CLIP~\cite{clip} and \DINO~\cite{dino} teachers, respectively.
\VQKDMAE{} and \VQKDViT{} represent tokenizers trained with \MAE~\cite{mae} and \ViT~\cite{vit} teachers.
The latter two teachers are pretrained on \INOneK{} utilizing a \ViTBSixteen{} architecture.

\subsection{Benchmark}
\label{sec:benchmark}

We detail how we fairly compare different tokenizers for token-based \IG{} here.

For each tokenizer, we train a proposal network $\P$ and a decoder $\D$ to constitute an image generator.
In training, the tokenizer is frozen to ensure fairness.
Thus, $\P$ and $\D$ can be trained in parallel.
We follow \VQGAN~\cite{vqgan} to train the \AR{} proposal network and the decoder.
The \NAR{} proposal network is trained following \MAGE~\cite{mage}.
Implementation details can be found in \Cref{app:implementation_details}.

Our benchmark adopts various metrics to comprehensively evaluate the image tokenizers.
Given an image tokenizer, we assess the effectiveness of its encoding process by evaluating the codebook usage.
To assess the generative capabilities of the image tokenizers, we evaluate \IS~\cite{is} and \FID~\cite{fid} on the generated images $\tilde{\II}$.
We assess the reconstruction capabilities of the image tokenizers by reporting the reconstruction \FID{} (\rFID).
In addition, we present the \PPL{} scores to appraise the \AR{} modeling proficiency of the image tokenizers.
A low \PPL{} score implies that $\PAR$ easily models $\z$.
Details about the evaluation metrics can be found in \Cref{app:evaluation}.

\subsection{Main Observation}
\label{sec:main_observation}

\begin{table}[t]
\centering
\caption{Comparison between image tokenizers on \INOneK.}
\label{tab:main}
\begin{tabular}{@{}lccccccc@{}}
\toprule
\multicolumn{1}{@{}c}{\multirow{2}{*}{Tokenizer $\T$}} & \multirow{2}{*}{\begin{tabular}[c]{@{}c@{}}Codebook\\      Usage (\%)\end{tabular}} & \multirow{2}{*}{\rFID{} $\downarrow$} & \multicolumn{3}{c}{$\PAR$}                                                                & \multicolumn{2}{c}{$\PNAR$}                                     \\ \cmidrule(lr){4-6}\cmidrule(l){7-8}
\multicolumn{1}{c}{}                                &                                                                                     &                                       & \PPL{} $\downarrow$       & \FID\textsubscript{\AR} $\downarrow$ & \IS\textsubscript{\AR} & \FID\textsubscript{\NAR} $\downarrow$ & \IS\textsubscript{\NAR} \\ \midrule
\VQGAN                                              & \phantom{00}4.9                                                                     & 5.09                                  & 116.75                    & 24.11                                & \phantom{0}39.52       & 20.03                                 & \phantom{0}48.30        \\
\FSQ                                                & \textbf{100.0}                                                                      & 4.96                                  & 791.56                    & 40.17                                & \phantom{0}26.40       & 29.78                                 & \phantom{0}33.63        \\ \midrule
\VQKDCLIP                                           & \textbf{100.0}                                                                      & 4.96                                  & \textbf{\phantom{0}53.73} & 11.78                                & \textbf{128.18}        & \phantom{0}9.51                       & \textbf{121.33}         \\
\VQKDViT                                            & \textbf{100.0}                                                                      & 3.69                                  & \phantom{0}89.30          & \textbf{11.40}                       & 107.56                 & \textbf{\phantom{0}8.45}              & 108.75                  \\
\VQKDDINO                                           & \textbf{100.0}                                                                      & \textbf{3.41}                         & \phantom{0}74.07          & 13.15                                & \phantom{0}80.89       & 10.21                                 & \phantom{0}91.39        \\
\VQKDMAE                                            & \textbf{100.0}                                                                      & 4.93                                  & 280.06                    & 26.85                                & \phantom{0}40.03       & 16.11                                 & \phantom{0}59.05        \\ \bottomrule
\end{tabular}
\end{table}

We evaluate the \textit{class-conditional \IG} performance of \VQGAN, \FSQ, and \VQKD{} tokenizers on \INOneK.
The results in \Cref{tab:main} leads to the following observations.

\textbf{\VQKD{} significantly enhances generation quality over \VQGAN.}
Equipped with either \AR{} or \NAR{} proposal networks, \VQKD{} tokenizers consistently outperform \VQGAN{} and \FSQ{}, as evidenced by superior \FID{} and \IS{} metrics.
In particular, \VQKDViT{} attains an \FIDAR{} of $11.40$ and an \FIDNAR{} of $8.45$, both less than half of those from \VQGAN{} ($24.11$ \FIDAR{} and $20.03$ \FIDNAR).

\Cref{tab:system} presents a system-level comparison between \VQKDCLIP{} and other class-conditional \IG{} models on \INOneK{} at a resolution of $256\times 256$. With a 1.4B \AR{} proposal network, \VQKDCLIP{} achieves an \FID{} of $4.10$, surpassing prior \AR, \NAR, and several \VQ-based diffusion models.


\begin{wraptable}{l}{0.57\textwidth}
  \vspace{12pt}
\caption{System level comparison on \INOneK.}
\label{tab:system}
\resizebox{\linewidth}{!}{
  \begin{tabular}{@{}cccc@{}}
  \toprule
  Model                          & Architecture & \#params & FID $\downarrow$         \\ \midrule
  \VQGAN~\cite{vqgan}            & \AR          & 1.4B     & 15.78                    \\
  \RQVAE~\cite{rq_vae}           & \AR          & 1.4B     & \phantom{0}8.71          \\
  \ViTVQGAN~\cite{vit_vqgan}     & \AR          & 1.7B     & \phantom{0}5.30          \\ \midrule
  \MoVQ~\cite{movq}              & \NAR         & 307M     & \phantom{0}8.78          \\
  \MaskGIT~\cite{maskgit}        & \NAR         & 227M     & \phantom{0}6.18          \\
  \FSQ~\cite{fsq}                & \NAR         & 227M     & \phantom{0}4.53          \\ \midrule
  LDM-8-G~\cite{ldm}             & Diffusion    & 506M     & \phantom{0}7.76          \\
  \CVQVAE~\cite{adm}             & Diffusion    & 400M     & \phantom{0}6.87          \\ \midrule
  \VQKDCLIP~\cite{beitv2} (ours) & \AR          & 1.4B     & \textbf{\phantom{0}4.10} \\ \bottomrule
  \end{tabular}
}
\end{wraptable}

\textbf{The superiority of \VQKD{} is irrelevant to the quantization operation and codebook usage.}
Both \VQKDCLIP{} and \FSQ{} record $100.0\%$ codebook usage and $4.96$ \rFID{}, but \VQKDCLIP{} achieves lower \FIDAR{} and higher \ISAR{} scores.
Moreover, \VQKD{} proves robustness to high codebook usage, with the \PPL{} metric of most \VQKD{} tokenizers surpassing \VQGAN.
In contrast, \FSQ{} lags behind \VQGAN{} in terms of \PPL, suggesting that the high codebook usage of \FSQ{} hinders $\PAR$ from modeling the code sequence $\z$.
As demonstrated in \Cref{sec:codebook_visualization}, this difference is likely due to the rich semantics in the \VQKD{} feature map.

\textbf{Tokenizers with stronger semantic understanding tend to deliver superior \IG{} performance.}
Considering the \FID{} and \IS{} metrics, we find that \VQKD{} tokenizers with supervised teachers (\CLIP{} and \ViT) consistently surpass those with unsupervised teachers (\DINO{} and \MAE).
While \VQKDDINO{} achieves the lowest \rFID{} and \PPL, its $13.15$ \FIDAR{} is worse than \VQKDCLIP{} ($11.78$) and \VQKDViT{} ($11.40$).
This trend can be attributed to the superior capability of supervised models in capturing semantics compared to the unsupervised ones.

\subsection{Further Verification}
\label{sec:further_verification}

\textbf{The superiority of \VQKD{} holds across proposal networks.}
As seen in \Cref{tab:main}, all \VQKD{} tokenizers surpass \VQGAN{} and \FSQ{} in the \FIDNAR{} and \ISNAR{} metrics.
In particular, \VQKDViT{} scores the lowest \FIDNAR{} at $8.45$ and \VQKDCLIP{} scores the highest \ISNAR{} at $121.33$.
In contrast, \VQGAN{} only achieves an \FIDNAR{} of $20.03$ and an \ISNAR{} of $48.30$.
It is also worth mentioning that \VQKDCLIP{} and \VQKDViT{} show slightly better performance than \VQKDDINO{} and \VQKDMAE{}, further supporting our conclusion that superior semantic understanding in supervised models plays a significant role in enhancing the quality of \IG.

\begin{table}[ht]
\centering
\caption{Comparison between image tokenizers on \MSCOCO{}. \TTwoI{} experiments are conducted on the \MSCOCO~Captions dataset.}
\label{tab:coco}
\begin{tabular}{@{}lccccc@{}}
\toprule
\multicolumn{1}{@{}c}{\multirow{2}{*}{Tokenizer $\T$}} & \multirow{2}{*}{\begin{tabular}[c]{@{}c@{}}Codebook\\      Usage (\%)\end{tabular}} & \multirow{2}{*}{\rFID{} $\downarrow$} & \multicolumn{2}{c}{$\PAR$}                                        & \multirow{2}{*}{\FID\textsubscript{\TTwoI} $\downarrow$} \\ \cmidrule(lr){4-5}
\multicolumn{1}{c}{}                                &                                                                                     &                                       & \PPL{} $\downarrow$        & \FID\textsubscript{\AR} $\downarrow$ &                                                          \\ \midrule
\VQGAN                                              & \phantom{00}2.4                                                                     & 16.21                                 & \textbf{\phantom{00}47.89} & 38.43                                & 24.11                                                    \\
\FSQ                                                & \textbf{100.0}                                                                      & \phantom{0}4.62                       & 1040.02                    & 44.64                                & 23.36                                                    \\ \midrule
\VQKDCLIP                                           & \phantom{0}82.2                                                                     & \phantom{0}5.48                       & \phantom{00}72.31          & 29.80                                & \textbf{11.17}                                           \\
\VQKDViT                                            & \textbf{100.0}                                                                      & \phantom{0}3.70                       & \phantom{0}117.10          & 23.51                                & 15.49                                                    \\
\VQKDDINO                                           & \textbf{100.0}                                                                      & \textbf{\phantom{0}2.69}              & \phantom{0}129.93          & \textbf{17.55}                       & 11.50                                                    \\
\VQKDMAE                                            & \textbf{100.0}                                                                      & \phantom{0}3.51                       & \phantom{0}317.98          & 44.01                                & 15.60                                                    \\ \bottomrule
\end{tabular}
\end{table}

\textbf{The superiority of \VQKD{} holds across datasets.}
We conduct \textit{unconditional \IG} experiments on the \MSCOCO{} dataset~\cite{coco}, which contains images of greater complexity than \INOneK.
As demonstrated in \Cref{tab:coco}, \VQKDDINO{} achieves a \FIDAR{} metric of $17.55$, significantly outperforming \VQGAN{} ($38.43$) and \FSQ{} ($44.64$).
Since the \ViT{} teacher in \VQKDViT{} is pretrained on \INOneK, the \rFID{} and \FIDAR{} metrics of \VQKDViT{} are slightly inferior to \VQKDDINO.
Note that the \PPL{} metric of \VQGAN{} is misleadingly favorable, due to its low codebook usage ($2.4\%$).

\textbf{The superiority of \VQKD{} holds across tasks.}
\textit{\TextToImage} (\TTwoI) experiments are conducted on the \MSCOCO{}~Captions dataset~\cite{coco_captions}.
As shown in \Cref{tab:coco}, the \FIDTTwoI{} metric of \VQKD{} tokenizers range from $11.17$ to $15.60$, while \VQGAN{} and \FSQ{} only achieves $24.11$ and $23.36$, respectively.

\section{Analysis}

In this section, we analyze tokenizers based on feature reconstruction from various perspectives.

\subsection{Codebook Visualization}
\label{sec:codebook_visualization}

\begin{figure}[ht]
  \centering
  \includegraphics[width=\linewidth]{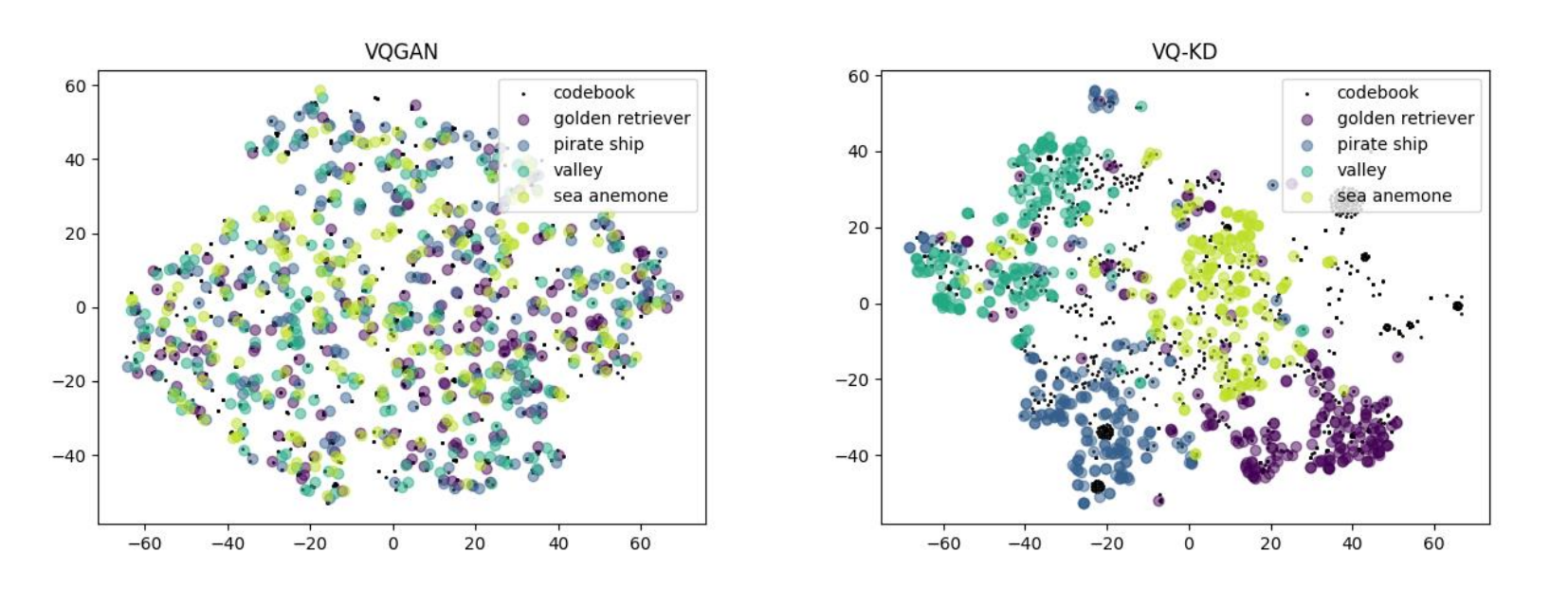}
  \caption{
    Codebook visualization of \VQGAN{} and \VQKDViT. 
    Best viewed in color.
  }
  \label{fig:tsne}
\end{figure}

We delve into the superiority of \VQKD{} by visualizing its codebook. 
From \INOneK, we randomly choose four categories: golden retriever, pirate ship, valley, and sea anemone. 
For each image belonging to these categories, we deploy \tSNE~\cite{tsne} to project the feature map $x$ and the code vectors $\mathbf{C}(\mathbf{z})$ into a two-dimensional space. 
$x$ is colored according to the image category and $\mathbf{C}(\mathbf{z})$ is illustrated in black.
As depicted in \Cref{fig:tsne}, the feature space of \VQKD{} shows superior organization compared to \VQGAN. 
In the \VQKD{} feature space, $x$ from the same category are grouped together. 
This implies that each code in the \VQKD{} codebook conveys clear semantics. 
As codes with similar semantics are likely to present concurrently in an image, it becomes easier for the proposal network to model the code sequence $\z$. 
Conversely, each code in the \VQGAN{} codebook is shared by multiple categories, resulting in semantics ambiguity. 
Hence, as illustrated in \Cref{tab:main}, the \PPL{} metric for \VQGAN{} is higher than \VQKD, even though its codebook usage is considerably lower.

\subsection{Clustering Pretrained Models as Tokenizers}
\label{sec:clustering_pretrained_models_as_tokenizers}

To better harness the semantics in \IU{} encoders, we propose a straightforward pipeline that transforms \IU{} encoders into tokenizers via feature clustering.
Given a pretrained \IU{} model $\T^{'}$, we employ it to encode the feature map $x^\T$ and subsequently utilize a clustering approach~\cite{cvq_vae} to acquire $N$ clusters.
The cluster centroids constitute a codebook $\C$.
$\T^{'}$ remains frozen during training, which significantly accelerates the training process.
As shown in \Cref{tab:cluster}, \ClusterViT{} presents $13.40$ \FIDAR{}, $10.58$ \FIDTTwoI, and $0.245$ \CLIP{} score on \MSCOCO{}, outperforming all tokenizers in \Cref{tab:coco}.
This suggests that pretrained models with simple feature clustering can become good tokenizers.
However, the cluster-based tokenizers behave worse in terms of \rFID, since they encode little appearance detail in $x^\T$, which is essential for exact reconstruction.
As a result, their \FID{} and \IS{} metrics on \INOneK{} are marginally weaker than those of their \VQKD{} counterparts.

\begin{table}[ht]
\centering
\caption{Performance of cluster-based tokenizers.}
\label{tab:cluster}
\begin{tabular}{@{}lcccccc@{}}
\toprule
\multicolumn{1}{@{}c}{\multirow{2}{*}{Encoder $\E$}} & \multicolumn{3}{c}{\INOneK}                                                                             & \multicolumn{3}{c}{\MSCOCO}                                                                               \\ \cmidrule(lr){2-4}\cmidrule(l){5-7}
\multicolumn{1}{c}{}                              & \rFID{} $\downarrow$     & \FID\textsubscript{\AR} $\downarrow$ & \FID\textsubscript{\NAR} $\downarrow$ & \rFID{} $\downarrow$     & \FID\textsubscript{\AR} $\downarrow$ & \FID\textsubscript{\TTwoI} $\downarrow$ \\ \midrule
\CLIP                                             & \phantom{0}9.13          & 14.81                                & 12.83                                 & \phantom{0}7.28          & 20.03                                & 12.82                                   \\
\ViT                                              & \textbf{\phantom{0}4.78} & \textbf{11.87}                       & \textbf{\phantom{0}8.59}              & \phantom{0}4.59          & \textbf{13.40}                       & 10.58                                   \\
\DINO                                             & \phantom{0}5.16          & 14.53                                & 11.23                                 & \textbf{\phantom{0}4.02} & 25.35                                & \textbf{\phantom{0}7.66}                \\
\MAE                                              & 15.15                    & 38.72                                & 34.26                                 & 10.08                    & 62.17                                & 18.85                                   \\ \bottomrule
\end{tabular}
\end{table}

\subsection{Scaling Up the Proposal Network}
\label{sec:scaling_up_the_proposal_network}

\begin{wraptable}{r}{0.55\textwidth}
\centering
\caption{\AR{} modeling with a large-scale proposal network or strong data augmentation.}
\label{tab:ar_modeling}
\begin{tabular}{@{}lcccc@{}}
\toprule
\multicolumn{1}{@{}c}{\multirow{2}{*}{Tokenizer $\T$}} & \multicolumn{2}{c}{\GPTTwoXL}                                   & \multicolumn{2}{c}{Strong Aug.}                               \\ \cmidrule(lr){2-3}\cmidrule(l){4-5}
\multicolumn{1}{c}{}                                & \FID\textsubscript{\AR}   $\downarrow$ & \IS\textsubscript{\AR} & \FID\textsubscript{\AR} $\downarrow$ & \IS\textsubscript{\AR} \\ \midrule
\VQGAN                                              & 17.13                                  & \phantom{0}59.19       & 32.76                                & 26.76                  \\
\FSQ                                                & 25.87                                  & \phantom{0}49.59       & 52.17                                & 18.09                  \\ \midrule
\VQKDCLIP                                           & 11.27                                  & \textbf{150.63}        & 15.24                                & \textbf{90.41}         \\
\VQKDViT                                            & \textbf{\phantom{0}9.23}               & 146.00                 & \textbf{13.32}                       & 81.44                  \\
\VQKDDINO                                           & \phantom{0}9.50                        & 120.26                 & 19.39                                & 52.74                  \\
\VQKDMAE                                            & 17.11                                  & \phantom{0}69.03       & 36.63                                & 28.49                  \\ \midrule
\ClusterCLIP                                        & 14.00                                  & 110.26                 & 17.22                                & 83.72                  \\ \bottomrule
\end{tabular}
\end{wraptable}

We examine the \IG{} performance of tokenizers with a large-scale proposal network.
Following \VQGAN, we adopt \GPTTwoXL{} as $\PAR$, which comes with $1.4$B parameters.
In line with \Cref{tab:main}, \VQKDViT{} leads with $9.23$ \FIDAR{}, while \VQKDCLIP{} achieves the highest \IS{} metric at $150.63$.
Upon comparing with \Cref{tab:main}, tokenizers with stronger \IU{} capabilities exhibit less improvement in the \FIDAR{} metric.
For instance, \VQKDMAE{} improves significantly from $26.85$ to $17.11$, while \VQKDCLIP{} reveals a marginal enhancement from $11.78$ to $11.27$.
This suggests that a small-scale $\PAR$ is sufficient for tokenizers with strong \IU{} capabilities, whereas those with weaker \IU{} abilities benefit from a large-scale $\PAR$.

\subsection{Influence of Strong Data Augmentation}
\label{sec:influence_of_strong_data_augmentation}

We investigate the impact of strong data augmentation on the \AR{} modeling performance of tokenizers.
Specifically, we employed a strong random crop, where the crop scale ranges from $0.08$ to $1.0$, introducing greater variability into the training data.
As shown in \Cref{tab:ar_modeling}, all tokenizers exhibit worse \FIDAR{} metrics than their counterparts in \Cref{tab:main}.
Interestingly, tokenizers with stronger \IU{} capabilities demonstrate greater robustness to the strong data augmentation.
For instance, \VQKDViT{} experiences a minor increase in \FIDAR{} of just $1.92$ (from $11.40$ to $13.32$), whilst \VQKDMAE{} records a considerable leap of $9.78$ (from $26.85$ to $36.63$).

\subsection{Large Teacher Models in \VQKD}
\label{sec:large_teacher_models_in_vq_kd}

\begin{wraptable}{r}{0.55\textwidth}
  \vspace{6pt}
\centering
\caption{Effect of different teachers in \VQKD.}
\label{tab:open_clip}
\begin{tabular}{@{}lcccc@{}}
\toprule
\multicolumn{1}{@{}c}{\OpenCLIP} & \rFID{} $\downarrow$ & \PPL{} $\downarrow$ & \FID\textsubscript{\AR}   $\downarrow$ & \IS\textsubscript{\AR} \\ \midrule
\ViTLFourteen                 & 4.03                 & 80.56               & 10.31                                  & 146.21                 \\
\ViTHFourteen                 & \textbf{3.60}        & 97.32               & \phantom{0}9.64                        & \textbf{161.13}        \\
\ViTGFourteen                 & 3.80                 & \textbf{77.79}      & \textbf{\phantom{0}8.70}               & 152.71                 \\ \bottomrule
\end{tabular}
\end{wraptable}

We incorporate \OpenCLIP~\cite{openclip} models of varying sizes as teacher models to train the \VQKD{} tokenizers.
As illustrated in \Cref{tab:open_clip}, the \FIDAR{} metric sees a reduction from $10.31$ to $8.70$ when the size of the \OpenCLIP{} model escalates from \ViTLFourteen{} to \ViTGFourteen.
Given that larger \OpenCLIP{} models inherently possess stronger \IU{} capabilities, these findings further corroborate the superiority of image tokenizers with more potent \IU{} capabilities.

\begin{table}[ht]
  \centering
  \caption{
    Effect of codebook size and dimension. Experiments are conducted on \VQKDCLIP.
  }
  \begin{subtable}{.515\linewidth}
\centering
\caption{Codebook size.}
\label{tab:codebook_size}
\resizebox{\textwidth}{!}{
\begin{tabular}{@{}ccccc@{}}
\toprule
\multicolumn{2}{@{}c}{Codebook   $\C$}           & \multirow{2}{*}{\rFID{} $\downarrow$} & \multirow{2}{*}{\FID\textsubscript{\AR} $\downarrow$} & \multirow{2}{*}{\IS\textsubscript{\AR}} \\ \cmidrule(r){1-2}
Size (log\textsubscript{2}) & Usage (\%)      &                                       &                                                       &                                         \\ \midrule
10                          & \textbf{100.0}  & 6.59                                  & 11.65                                                 & 114.90                                  \\
11                          & \textbf{100.0}  & 5.99                                  & \textbf{10.98}                                        & 119.72                                  \\
12                          & \textbf{100.0}  & 5.64                                  & 11.71                                                 & 123.42                                  \\
13                          & \textbf{100.0}  & 4.96                                  & 11.78                                                 & 128.18                                  \\
14                          & \phantom{0}93.7 & \textbf{4.53}                         & 11.61                                                 & \textbf{131.51}                         \\ \bottomrule
\end{tabular}
}
  \end{subtable}
  \hfill
  \begin{subtable}{.46\linewidth}
\centering
\caption{Codebook dimension.}
\label{tab:codebook_dimension}
\resizebox{\textwidth}{!}{
\begin{tabular}{@{}ccccc@{}}
\toprule
\multicolumn{2}{@{}c}{Codebook   $\C$} & \multirow{2}{*}{\rFID{} $\downarrow$} & \multirow{2}{*}{\FID\textsubscript{\AR} $\downarrow$} & \multirow{2}{*}{\IS\textsubscript{\AR}} \\ \cmidrule(r){1-2}
Dim        & Usage (\%)             &                                       &                                                       &                                         \\ \midrule
16         & \textbf{100.0}         & 4.94                                  & 12.19                                                 & 124.71                                  \\
32         & \textbf{100.0}         & 4.96                                  & 11.78                                                 & \textbf{128.18}                         \\
64         & \phantom{0}97.4        & \textbf{4.64}                         & 11.00                                                 & 126.60                                  \\
128        & \phantom{0}89.6        & 5.03                                  & \textbf{10.50}                                        & 119.78                                  \\
256        & \phantom{0}48.7        & 6.80                                  & 12.08                                                 & 103.07                                  \\ \bottomrule
\end{tabular}
}
  \end{subtable}
\end{table}

\begin{figure}[ht]
  \centering
  \includegraphics[width=\linewidth]{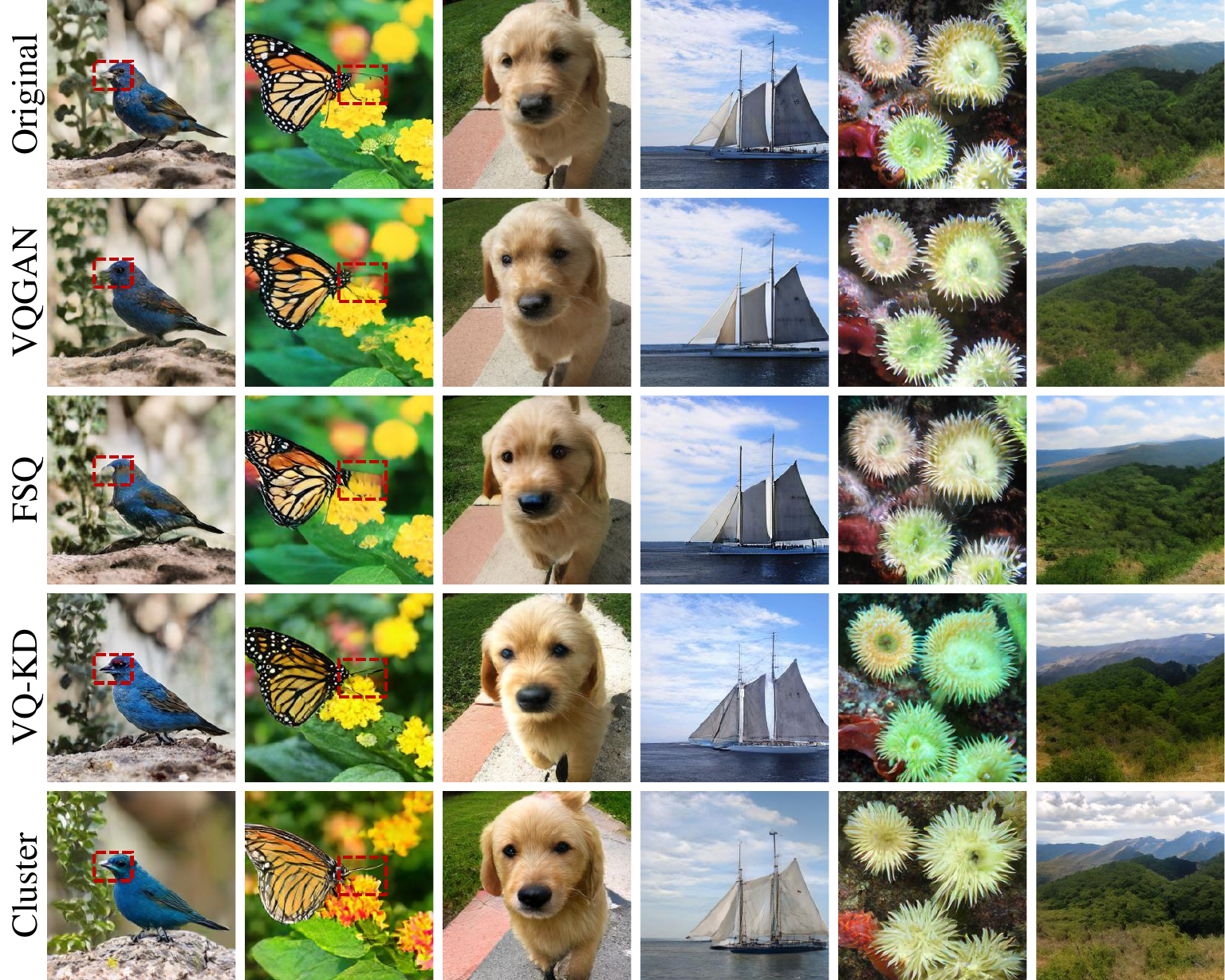}
  \caption{
    Reconstruction results of different image tokenizers.
  }
  \label{fig:reconstruction}
\end{figure}

\subsection{Codebook Size and Dimension}
\label{sec:codebook_size_and_dimension}

The size and dimension of the codebook exert a significant influence on the \IG{} performance of tokenizers~\cite{vqgan, vit_vqgan}.
\Cref{tab:codebook_size} showcases the performance of \VQKDCLIP{} with varying codebook sizes.
Large codebooks aid the tokenizers in representing fine-grained semantics, contributing to a consistent decrease in the \rFID{} metric from $6.59$ to $4.53$.
The \IS{} metric also shows favor towards larger codebooks, with size $2^{14}$ leading to the highest \IS{} metric of $131.51$.
However, choosing the correct code from a large codebook is harder than from a small codebook, hindering $\PAR$ from achieving lower \FID{} scores with larger codebooks.

\Cref{tab:codebook_dimension} demonstrates the influence of codebook dimension.
High-dimensional codes carry more information but lead to lower codebook usage.
As a result, the \rFID{} metric initially drops from $4.96$ to $4.64$, then increases drastically to $6.80$.
Similar to \Cref{tab:codebook_size}, the \FID{} and \IS{} metrics favor different codebook dimensions.
\FID{} favors $128$-dimensional codebooks, where codebook usage is relatively low.
In contrast, \IS{} favors $32$-dimensional codebooks, possibly due to a superior diversity.


\subsection{Qualitative Analysis}
\label{sec:quanlitative_analysis}


\begin{figure}[ht]
  \centering
  \includegraphics[width=0.93\linewidth]{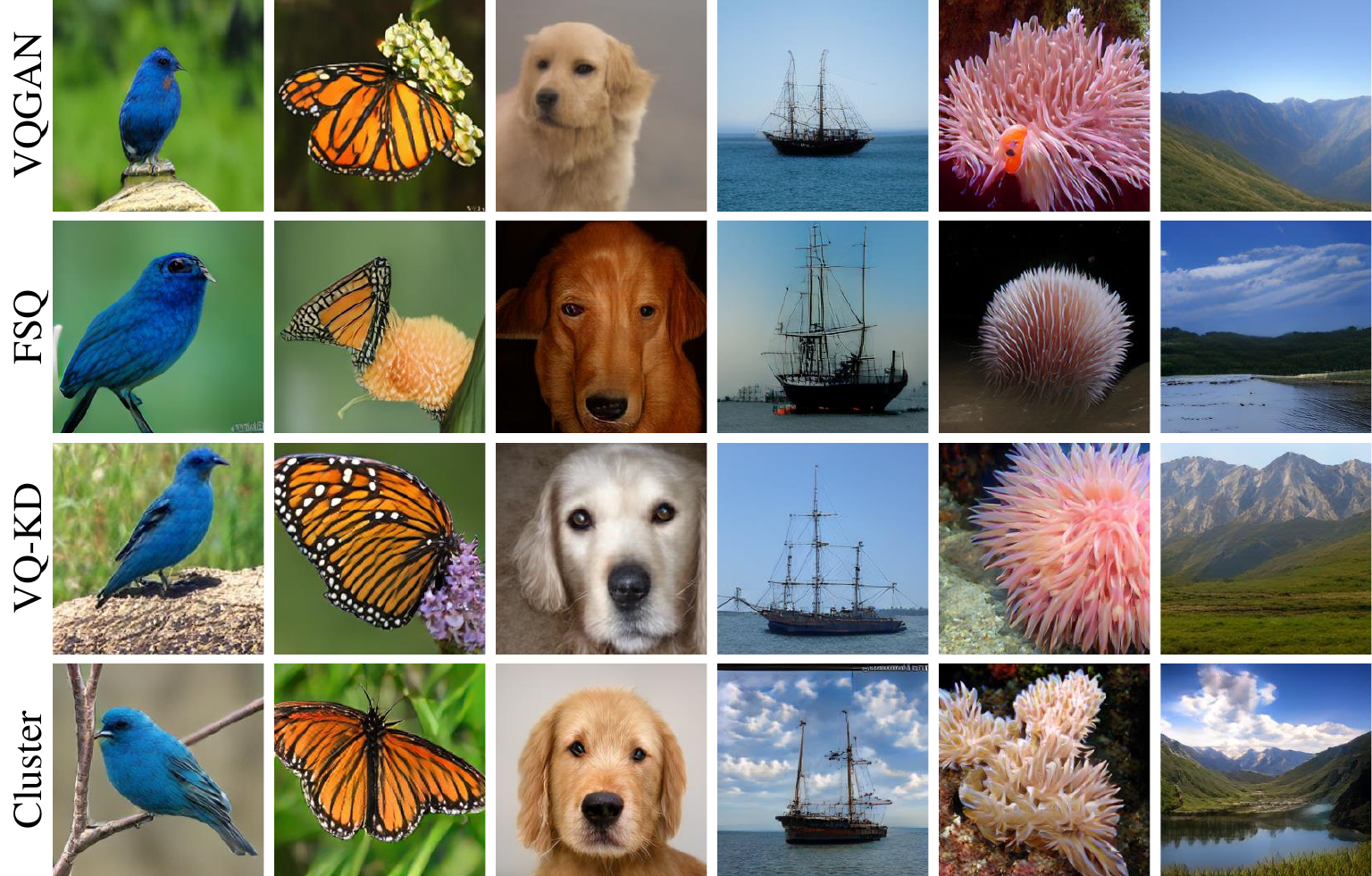}
  \caption{
    Class-conditional \AR{} generation results of different image tokenizers.
  }
  \label{fig:ar}
\end{figure}

\begin{figure}[ht]
  \centering
  \includegraphics[width=0.93\linewidth]{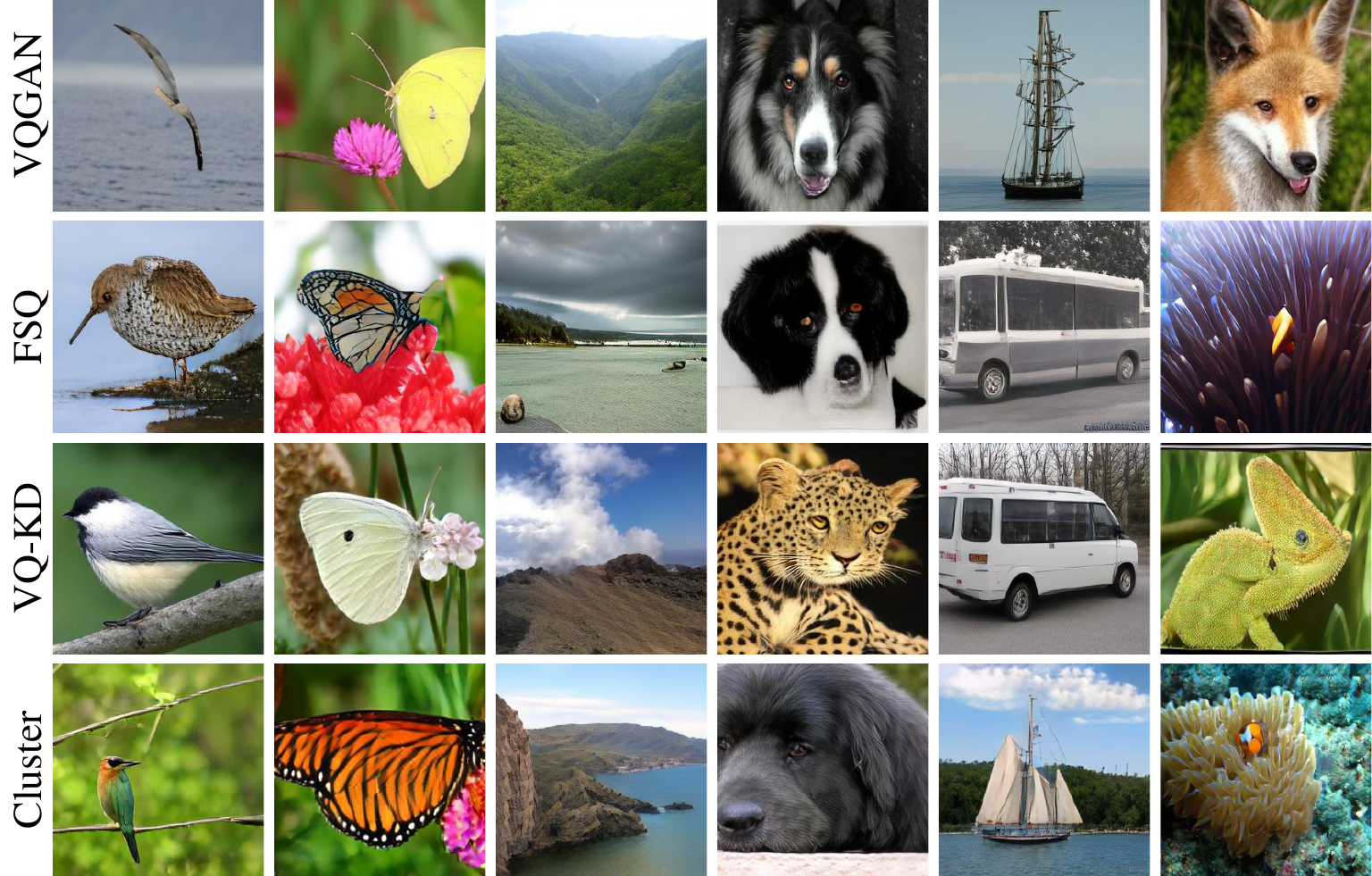}
  \caption{
    Unconditional \NAR{} generation results of different image tokenizers.
  }
  \label{fig:nar}
\end{figure}

The reconstruction quality of various tokenizers is demonstrated in \Cref{fig:reconstruction}. 
Original images are displayed in the first row. 
Regions where \VQGAN{} and \FSQ{} fail to reconstruct are highlighted with red boxes. 
In contrast, \VQKD{} reconstructions are visually more accurate. 
Since the \IU{} encoder in \Cluster{} emphasizes encoding semantics over visual details, \Cluster{} fails to preserve all visual details during reconstruction.
Nonetheless, the reconstruction results of \Cluster{} still appear more natural than \VQGAN and \FSQ, especially in the highlighted areas.
\Cref{fig:ar} and \Cref{fig:nar} further illustrates the \AR{} and \NAR{} generation results, showcasing the superior visual performance of \VQKD{} and \Cluster.

\section{Conclusion}

In this paper, we show that image understanding (\IU) models can be useful in image generation (\IG).
Through comprehensive studies, observe that the \VQKD{} tokenizers significantly enhance generation quality over \VQGAN, irrelevant of the quantization operation and codebook utilization.
Within the \VQKD{} tokenizers, stronger \IU{} capabilities tends to deliver superior \IG{} performance.
Further verification shows that the superiority of \VQKD{} holds across proposal networks, datasets, and tasks.
Lastly, we analyze \VQKD{} from multiple angles, including clustering pretrained models as tokenizers, scaling up the proposal network, influence of strong data augmentation, large teacher models in \VQKD, and codebook size and dimensions.

\newpage

\section*{Acknowledgement}

This research is supported in part by National Key R\&D Program of China (2022ZD0115502), National Natural Science Foundation of China (NO. 62122010, U23B2010), Zhejiang Provincial Natural Science Foundation of China under Grant No. LDT23F02022F02, Key Research and Development Program of Zhejiang Province under Grant 2022C01082, "Pioneer" and "Leading Goose" R\&D Program of Zhejiang (No. 2024C01161).


\begin{thebibliography}{10}

\bibitem{ste}
Yoshua Bengio, Nicholas L{\'{e}}onard, and Aaron Courville.
\newblock {Estimating or Propagating Gradients Through Stochastic Neurons for Conditional Computation}.
\newblock {\em arXiv preprint arXiv:1308.3432}, 2013.

\bibitem{gpt_3}
Tom~B. Brown, Benjamin Mann, Nick Ryder, Melanie Subbiah, Jared Kaplan, Prafulla Dhariwal, Arvind Neelakantan, Pranav Shyam, Girish Sastry, Amanda Askell, Sandhini Agarwal, Ariel Herbert-Voss, Gretchen Krueger, Tom Henighan, Rewon Child, Aditya Ramesh, Daniel~M. Ziegler, Jeffrey Wu, Clemens Winter, Christopher Hesse, Mark Chen, Eric Sigler, Mateusz Litwin, Scott Gray, Benjamin Chess, Jack Clark, Christopher Berner, Sam McCandlish, Alec Radford, Ilya Sutskever, and Dario Amodei.
\newblock {Language Models are Few-Shot Learners}.
\newblock In {\em NeurIPS}, 2020.

\bibitem{dino}
Mathilde Caron, Hugo Touvron, Ishan Misra, Herv{\'{e}} Jegou, Julien Mairal, Piotr Bojanowski, and Armand Joulin.
\newblock {Emerging Properties in Self-Supervised Vision Transformers}.
\newblock In {\em ICCV}, 2021.

\bibitem{muse}
Huiwen Chang, Han Zhang, Jarred Barber, AJ~Maschinot, Jose Lezama, Lu~Jiang, Ming-Hsuan Yang, Kevin Murphy, William~T. Freeman, Michael Rubinstein, Yuanzhen Li, and Dilip Krishnan.
\newblock {Muse: Text-To-Image Generation via Masked Generative Transformers}.
\newblock In {\em ICML}, 2023.

\bibitem{maskgit}
Huiwen Chang, Han Zhang, Lu~Jiang, Ce~Liu, and William~T Freeman.
\newblock {MaskGIT: Masked Generative Image Transformer}.
\newblock In {\em CVPR}, 2022.

\bibitem{coco_captions}
Xinlei Chen, Hao Fang, Tsung-yi Lin, Ramakrishna Vedantam, Saurabh Gupta, Piotr Dollar, and C~Lawrence Zitnick.
\newblock {Microsoft COCO Captions: Data Collection and Evaluation Server}.
\newblock {\em arXiv preprint arXiv:1504.00325}, 2015.

\bibitem{imagenet}
Jia Deng, Wei Dong, Richard Socher, Li-Jia Li, {Kai Li}, and {Li Fei-Fei}.
\newblock {ImageNet: A large-scale hierarchical image database}.
\newblock In {\em CVPR}, 2009.

\bibitem{adm}
Prafulla Dhariwal and Alex Nichol.
\newblock {Diffusion Models Beat GANs on Image Synthesis}.
\newblock In {\em NeurIPS}, volume~11, pages 8780--8794, 2021.

\bibitem{vit}
Alexey Dosovitskiy, Lucas Beyer, Alexander Kolesnikov, Dirk Weissenborn, Xiaohua Zhai, Thomas Unterthiner, Mostafa Dehghani, Matthias Minderer, Georg Heigold, Sylvain Gelly, Jakob Uszkoreit, and Neil Houlsby.
\newblock {An Image is Worth 16x16 Words: Transformers for Image Recognition at Scale}.
\newblock In {\em ICLR}, 2021.

\bibitem{vqgan}
Patrick Esser, Robin Rombach, and Bjorn Ommer.
\newblock {Taming Transformers for High-Resolution Image Synthesis}.
\newblock In {\em CVPR}, 2021.

\bibitem{vq}
R.~Gray.
\newblock {Vector quantization}.
\newblock {\em IEEE ASSP Mag.}, 1(2), 1984.

\bibitem{seq_gan}
Yuchao Gu, Xintao Wang, Yixiao Ge, Ying Shan, Xiaohu Qie, and Mike~Zheng Shou.
\newblock {Rethinking the Objectives of Vector-Quantized Tokenizers for Image Synthesis}.
\newblock {\em arXiv preprint arXiv:2212.03185}, 2022.

\bibitem{mae}
Kaiming He, Xinlei Chen, Saining Xie, Yanghao Li, Piotr Doll{\'{a}}r, and Ross Girshick.
\newblock {Masked Autoencoders Are Scalable Vision Learners}.
\newblock In {\em CVPR}, 2022.

\bibitem{fid}
Martin Heusel, Hubert Ramsauer, Thomas Unterthiner, Bernhard Nessler, and Sepp Hochreiter.
\newblock {GANs Trained by a Two Time-Scale Update Rule Converge to a Local Nash Equilibrium}.
\newblock In {\em NeurIPS}, 2017.

\bibitem{ddpm}
Jonathan Ho, Ajay Jain, and Pieter Abbeel.
\newblock {Denoising Diffusion Probabilistic Models}.
\newblock In {\em NeurIPS}, 2020.

\bibitem{soda}
Drew~A. Hudson, Daniel Zoran, Mateusz Malinowski, Andrew~K. Lampinen, Andrew Jaegle, James~L. McClelland, Loic Matthey, Felix Hill, and Alexander Lerchner.
\newblock {SODA: Bottleneck Diffusion Models for Representation Learning}.
\newblock {\em arXiv preprint arXiv:2311.17901}, 2023.

\bibitem{openclip}
Gabriel Ilharco, Mitchell Wortsman, Ross Wightman, Cade Gordon, Nicholas Carlini, Rohan Taori, Achal Dave, Vaishaal Shankar, Hongseok Namkoong, John Miller, Hannaneh Hajishirzi, Ali Farhadi, and Ludwig Schmidt.
\newblock Openclip.
\newblock \url{https://doi.org/10.5281/zenodo.5143773}, 2021.

\bibitem{adam}
Diederik~P. Kingma and Jimmy~Lei Ba.
\newblock {Adam: A method for stochastic optimization}.
\newblock In {\em ICLR}, 2015.

\bibitem{vae}
Diederik~P. Kingma and Max Welling.
\newblock {Auto-encoding variational bayes}.
\newblock {\em ICLR}, 2014.

\bibitem{videopoet}
Dan Kondratyuk, Lijun Yu, Xiuye Gu, Jos{\'{e}} Lezama, Jonathan Huang, Rachel Hornung, Hartwig Adam, Hassan Akbari, Yair Alon, Vighnesh Birodkar, Yong Cheng, Ming-Chang Chiu, Josh Dillon, Irfan Essa, Agrim Gupta, Meera Hahn, Anja Hauth, David Hendon, Alonso Martinez, David Minnen, David Ross, Grant Schindler, Mikhail Sirotenko, Kihyuk Sohn, Krishna Somandepalli, Huisheng Wang, Jimmy Yan, Ming-Hsuan Yang, Xuan Yang, Bryan Seybold, and Lu~Jiang.
\newblock {VideoPoet: A Large Language Model for Zero-Shot Video Generation}.
\newblock {\em arXiv preprint arXiv:2312.14125}, 2023.

\bibitem{rq_vae}
Doyup Lee, Chiheon Kim, Saehoon Kim, Minsu Cho, and Wook~Shin Han.
\newblock {Autoregressive Image Generation using Residual Quantization}.
\newblock In {\em CVPR}, 2022.

\bibitem{mage}
Tianhong Li, Huiwen Chang, Shlok~Kumar Mishra, Han Zhang, Dina Katabi, and Dilip Krishnan.
\newblock {MAGE: MAsked Generative Encoder to Unify Representation Learning and Image Synthesis}.
\newblock In {\em CVPR}, 2023.

\bibitem{coco}
Tsung-Yi Lin, Michael Maire, Serge Belongie, Lubomir Bourdev, Ross Girshick, James Hays, Pietro Perona, Deva Ramanan, C.~Lawrence Zitnick, and Piotr Doll{\'{a}}r.
\newblock {Microsoft COCO: Common Objects in Context}.
\newblock In {\em ECCV}, 2014.

\bibitem{adamw}
Ilya Loshchilov and Frank Hutter.
\newblock {Decoupled weight decay regularization}.
\newblock In {\em ICLR}, 2019.

\bibitem{fsq}
Fabian Mentzer, David Minnen, Eirikur Agustsson, and Michael Tschannen.
\newblock {Finite Scalar Quantization: VQ-VAE Made Simple}.
\newblock In {\em ICLR}, 2024.

\bibitem{gd}
Soumik Mukhopadhyay, Matthew Gwilliam, Vatsal Agarwal, Namitha Padmanabhan, Archana Swaminathan, Srinidhi Hegde, Tianyi Zhou, and Abhinav Shrivastava.
\newblock {Diffusion Models Beat GANs on Image Classification}.
\newblock {\em arXiv preprint arXiv:2307.08702}, 2023.

\bibitem{beitv2}
Zhiliang Peng, Li~Dong, Hangbo Bao, Qixiang Ye, and Furu Wei.
\newblock {BEiT v2: Masked Image Modeling with Vector-Quantized Visual Tokenizers}.
\newblock {\em arXiv preprint arXiv:2208.06366}, 2022.

\bibitem{diff_ae}
Konpat Preechakul, Nattanat Chatthee, Suttisak Wizadwongsa, and Supasorn Suwajanakorn.
\newblock {Diffusion Autoencoders: Toward a Meaningful and Decodable Representation}.
\newblock In {\em CVPR}, 2022.

\bibitem{clip}
Alec Radford, Jong~Wook Kim, Chris Hallacy, Aditya Ramesh, Gabriel Goh, Sandhini Agarwal, Girish Sastry, Amanda Askell, Pamela Mishkin, Jack Clark, Gretchen Krueger, and Ilya Sutskever.
\newblock {Learning Transferable Visual Models From Natural Language Supervision}.
\newblock In {\em ICML}, volume 139, 2021.

\bibitem{gpt}
Alec Radford, Karthik Narasimhan, Tim Salimans, and Ilya Sutskever.
\newblock {Improving Language Understanding by Generative Pre-Training}.
\newblock 2018.

\bibitem{gpt_2}
Alec Radford, Jeffrey Wu, Rewon Child, David Luan, Dario Amodei, and Ilya Sutskever.
\newblock {Language Models are Unsupervised Multitask Learners}.
\newblock {\em arXiv preprint arXiv:2007.07582}, 2019.

\bibitem{vq_vae_2}
Ali Razavi, A{\"{a}}ron van~den Oord, and Oriol Vinyals.
\newblock {Generating diverse high-fidelity images with VQ-VAE-2}.
\newblock In {\em NeurIPS}, 2019.

\bibitem{ldm}
Robin Rombach, Andreas Blattmann, Dominik Lorenz, Patrick Esser, and Bjorn Ommer.
\newblock {High-Resolution Image Synthesis with Latent Diffusion Models}.
\newblock In {\em CVPR}, 2022.

\bibitem{is}
Tim Salimans, Ian Goodfellow, Wojciech Zaremba, Vicki Cheung, Alec Radford, and Xi~Chen.
\newblock {Improved Techniques for Training GANs}.
\newblock In {\em NeurIPS}, 2016.

\bibitem{dpm}
Jascha Sohl-Dickstein, Eric~A. Weiss, Niru Maheswaranathan, and Surya Ganguli.
\newblock {Deep Unsupervised Learning using Nonequilibrium Thermodynamics}.
\newblock In {\em ICML}, 2015.

\bibitem{ddim}
Jiaming Song, Chenlin Meng, and Stefano Ermon.
\newblock {Denoising Diffusion Implicit Models}.
\newblock In {\em ICLR}, 2020.

\bibitem{ncsn}
Yang Song and Stefano Ermon.
\newblock {Generative Modeling by Estimating Gradients of the Data Distribution}.
\newblock In {\em NeurIPS}, 2019.

\bibitem{inception_v3}
Christian Szegedy, Vincent Vanhoucke, Sergey Ioffe, Jonathon Shlens, and Zbigniew Wojna.
\newblock {Rethinking the Inception Architecture for Computer Vision}.
\newblock In {\em CVPR}, 2016.

\bibitem{hq_vae}
Yuhta Takida, Yukara Ikemiya, Takashi Shibuya, Kazuki Shimada, Woosung Choi, Chieh-Hsin Lai, Naoki Murata, Toshimitsu Uesaka, Kengo Uchida, Wei-Hsiang Liao, and Yuki Mitsufuji.
\newblock {HQ-VAE: Hierarchical Discrete Representation Learning with Variational Bayes}.
\newblock {\em arXiv preprint arXiv:2401.00365}, 2023.

\bibitem{sq_vae}
Yuhta Takida, Takashi Shibuya, Wei~Hsiang Liao, Chieh~Hsin Lai, Junki Ohmura, Toshimitsu Uesaka, Naoki Murata, Shusuke Takahashi, Toshiyuki Kumakura, and Yuki Mitsufuji.
\newblock {SQ-VAE: Variational Bayes on Discrete Representation with Self-annealed Stochastic Quantization}.
\newblock {\em Proc. Mach. Learn. Res.}, 162, 2022.

\bibitem{vq_vae}
Aaron {Van Den Oord}, Oriol Vinyals, and Koray Kavukcuoglu.
\newblock {Neural discrete representation learning}.
\newblock In {\em NeurIPS}, 2017.

\bibitem{tsne}
Laurens {Van Der Maaten} and Geoffrey Hinton.
\newblock {Visualizing Data using t-SNE}.
\newblock {\em JMLR}, 9, 2008.

\bibitem{transformer}
Ashish Vaswani, Noam Shazeer, Niki Parmar, Jakob Uszkoreit, Llion Jones, Aidan~N. Gomez, Lukasz Kaiser, and Illia Polosukhin.
\newblock {Attention Is All You Need}.
\newblock In {\em NeurIPS}, 2017.

\bibitem{vq_wae}
Tung-Long Vuong, Trung Le, He~Zhao, Chuanxia Zheng, Mehrtash Harandi, Jianfei Cai, and Dinh Phung.
\newblock {Vector Quantized Wasserstein Auto-Encoder}.
\newblock {\em ICML}, 2023.

\bibitem{datasetdm}
Weijia Wu, Yuzhong Zhao, Hao Chen, Yuchao Gu, Rui Zhao, Yefei He, Hong Zhou, Mike~Zheng Shou, and Chunhua Shen.
\newblock {DatasetDM: Synthesizing Data with Perception Annotations Using Diffusion Models}.
\newblock In {\em NeurIPS}, 2023.

\bibitem{freemask}
Lihe Yang, Xiaogang Xu, Bingyi Kang, Yinghuan Shi, and Hengshuang Zhao.
\newblock {FreeMask: Synthetic Images with Dense Annotations Make Stronger Segmentation Models}.
\newblock In {\em NeurIPS}, 2023.

\bibitem{vit_vqgan}
Jiahui Yu, Xin Li, Jing~Yu Koh, Han Zhang, Ruoming Pang, James Qin, Alexander Ku, Yuanzhong Xu, Jason Baldridge, and Yonghui Wu.
\newblock {Vector-Quantized Image Modeling With Improved Vqgan}.
\newblock In {\em ICLR}, 2022.

\bibitem{parti}
Jiahui Yu, Yuanzhong Xu, Jing~Yu Koh, Thang Luong, Gunjan Baid, Zirui Wang, Vijay Vasudevan, Alexander Ku, Yinfei Yang, Burcu~Karagol Ayan, Ben Hutchinson, Wei Han, Zarana Parekh, Xin Li, Han Zhang, Jason Baldridge, and Yonghui Wu.
\newblock {Scaling Autoregressive Models for Content-Rich Text-to-Image Generation}.
\newblock {\em TMLR}, 2022.

\bibitem{magvitv2}
Lijun Yu, Jos{\'{e}} Lezama, Nitesh~B. Gundavarapu, Luca Versari, Kihyuk Sohn, David Minnen, Yong Cheng, Agrim Gupta, Xiuye Gu, Alexander~G. Hauptmann, Boqing Gong, Ming-Hsuan Yang, Irfan Essa, David~A. Ross, and Lu~Jiang.
\newblock {Language Model Beats Diffusion -- Tokenizer is Key to Visual Generation}.
\newblock {\em arXiv preprint arXiv:2310.05737}, 2023.

\bibitem{cm3leon}
Lili Yu, Bowen Shi, Ramakanth Pasunuru, Benjamin Muller, Olga Golovneva, Tianlu Wang, Arun Babu, Binh Tang, Brian Karrer, Shelly Sheynin, Candace Ross, Adam Polyak, Russell Howes, Vasu Sharma, Puxin Xu, Hovhannes Tamoyan, Oron Ashual, Uriel Singer, Shang-Wen Li, Susan Zhang, Gargi Ghosh, Yaniv Taigman, Maryam Fazel-Zarandi, Asli Celikyilmaz, Luke Zettlemoyer, and Armen Aghajanyan.
\newblock {Scaling Autoregressive Multi-Modal Models: Pretraining and Instruction Tuning}.
\newblock {\em arXiv preprint arXiv:2309.02591}, 2023.

\bibitem{diffusionengine}
Manlin Zhang, Jie Wu, Yuxi Ren, Ming Li, Jie Qin, Xuefeng Xiao, Wei Liu, Rui Wang, Min Zheng, and Andy~J. Ma.
\newblock {DiffusionEngine: Diffusion Model is Scalable Data Engine for Object Detection}.
\newblock {\em arXiv preprint arXiv:2309.03893}, 2023.

\bibitem{pdae}
Zijian Zhang, Zhou Zhao, and Zhijie Lin.
\newblock {Unsupervised Representation Learning from Pre-trained Diffusion Probabilistic Models}.
\newblock In {\em NeurIPS}, volume~35, 2022.

\bibitem{vpd}
Wenliang Zhao, Yongming Rao, Zuyan Liu, Benlin Liu, Jie Zhou, and Jiwen Lu.
\newblock {Unleashing Text-to-Image Diffusion Models for Visual Perception}.
\newblock In {\em ICCV}, 2023.

\bibitem{movq}
Chuanxia Zheng, Jianfei Cai, Long~Tung Vuong, and Dinh Phung.
\newblock {MoVQ: Modulating Quantized Vectors for High-Fidelity Image Generation}.
\newblock In {\em NeurIPS}, 2022.

\bibitem{cvq_vae}
Chuanxia Zheng and Andrea Vedaldi.
\newblock {Online Clustered Codebook}.
\newblock In {\em CVPR}, 2023.

\end{thebibliography}

\newpage
\appendix




\section{Datasets}
\label{app:datasets}

The experiments are conducted on two image datasets: \ImageNetOneK{} (\INOneK)~\cite{imagenet} and \MSCOCO~\cite{coco}.
The \INOneK{} dataset contains approximately $1.28$ million training images and $50,000$ validation images across $1,000$ diverse categories.
The \MSCOCO{} dataset comprises $82,783$ images for training and $40,504$ for validation. 
Each image is annotated with several captions.

For a given image, we first resize its shorter side to $s$ pixels, where $s$ symbolizes the input size.
Subsequently, a central crop is performed to derive an image fragment sized $s \times s$ pixels.
Our default augmentation strategy incorporates a random crop (between 0.8 and 1.0) partnered with random horizontal flipping.

\section{Implementation Details}
\label{app:implementation_details}

The image tokenizers under consideration generate token sequences of length $256$ upon a $256 \times 256$ input image.
All tokenizers remain frozen throughout the training of the decoder and the proposal networks.
Experiments are performed using $8$ A100 $80$GB GPUs.
The approximate training times for the \VQKD{} tokenizer is around $30$ hours, the decoder requires roughly $68$ hours, while the \AR{} proposal network and \NAR{} proposal network necessitate about $29$ hours and $72$ hours, respectively.
In total, a single experiment takes approximately $200$ hours training.

The CNN-based \VQGAN{} tokenizers, with $27.9$M parameters, are trained following the identical procedure employed for decoders.
The codebook dimension of \VQGAN{} is $256$.
\FSQ{} levels $\mathcal{L}$ are set to $(8, 8, 5, 5, 5)$, equivalent to codebook size $8,000$.

As per \BEiTvTwo~\cite{beitv2}, an \AdamW{} optimizer is utilized to train the \VQKD{} tokenizers.
The learning rate warms up linearly to $10^{-4}$ for $25,000$ steps, subsequently decaying to $10^{-5}$ under a cosine schedule.
Unless specifically stated, \VQKD{} tokenizer is trained with an input size of $224 \times 224$ and codebook dimension of $32$.

\VQGAN~\cite{vqgan} is adopted for training both the decoder and the AR proposal networks.
Both $\D$ and $\PAR$ training span $260,000$ steps with a collective batch size of $96$ for \INOneK{} and $24$ for \MSCOCO.
The decoder is a CNN-based \VQGAN{} decoder, consisting of $40.5$M parameters.
The decoders utilize the \Adam~\cite{adam} optimizer with learning rates set at $5.4 \times 10^{-5}$, $\beta_1 = 0.5$, and $\beta_2 = 0.9$.
Their discriminators are also trained via \Adam{} optimizer, employing learning rates of $4.32 \times 10^{-4}$, while keeping the $\beta$ constants identical.
Subsequent training of \AR{} proposal networks relies on the \AdamW~\cite{adamw} optimizer with $\beta_1 = 0.9$, $\beta_2 = 0.98$, and a $0.2$ weight decay.
An initial learning rate of $10^{-4}$ is set, after which it decays to $0$ on a cosine schedule.
The \AR{} proposal network is a \GPTTwoMedium~\cite{gpt_2}, with $335$M parameters.

We follow \MAGE~\cite{mage} for training \NAR{} proposal networks.
$\PNAR$ is trained for $300$ epochs with a collective batch size of $512$ on \ImageNetOneK.
\NAR{} proposal networks are trained with the \AdamW{} optimizer with $\beta_1 = 0.9$, $\beta_2 = 0.95$, and a $0.05$ weight decay.
The learning rate warms up linearly to $3 \times 10^{-4}$ throughout $10$ epochs before decaying to $0$ following a cosine schedule.
The encoder of $\PNAR$ is a \ViTBSixteen, with $86$M parameters.

\section{Evaluation}
\label{app:evaluation}

Codebook usage is defined as the proportion of codes from the codebook that have been used at least once when encoding the dataset.
A low value for codebook usage might be an indication of the `codebook collapse' issue.


\IS{} provides a measure of both the fidelity and diversity of $\tilde{\II}$.
However, \IS{} significantly relies on the classification capabilities of a pretrained \InceptionVThree{} model~\cite{inception_v3}.
Complex images are likely to be misinterpreted as lacking fidelity by \IS.
Therefore, we limit the use of \IS{} to \INOneK{} experiments only.

To circumvent the limitations of \IS, \FID{} computes the statistical distance in the \InceptionVThree{} feature space between the real images $\II$ and the generated images $\tilde{\II}$.
A lower \FID{} score indicates that $\tilde{\II}$ is statistically similar to $\II$.

\rFID{} is defined as the \FID{} between $\II$ and their reconstructed counterparts $\hat{\II}$.
Obtaining a low \rFID{} score requires that the image tokenizer encode sufficient visual details within the codes $\C(\z)$ to enable accurate reconstruction by the decoder.

The \PPL{} score is defined as:
\begin{equation}
  \text{\PPL} = \exp\left( -\frac1L \sum_{i = 1}^L \log p(z_i|z_{1:i-1}) \right),
\end{equation}
where $\z$ denotes a sequence of codes offered by the tokenizer, $L$ represents the length of $\z$, and $p(\z)$ embodies the distribution modeled by the \AR{} proposal network $\PAR$.

Both reconstruction and \AR{} modeling serve as two pivotal capabilities in an image generator.
We anticipate that these metrics will lead to a more thorough insight into the generative capacities of image tokenizers.





\section{Limitations}
\label{app:limitations}

The \VQKD{} tokenizers are designed to mimic the \IU{} encoders, yielding superior quantitative results compared to traditional tokenizers like the \VQGAN{} and \FSQ. 
Nonetheless, qualitative analysis suggest that the \VQKD{} may modify visual details during the pixel reformation process, thereby posing challenges for tasks such as image editing.

\section{Broader Impacts}
\label{app:broader_impacts}

This paper explores the question \textit{how might image understanding (\IU) models aid image generation (\IG) tasks}.
We envision that our findings will motivate research on image tokenizers and prompt the community to reconsider the correlation between \IU{} and \IG.

\newpage
\section*{NeurIPS Paper Checklist}

\begin{enumerate}
  \item {\bf Claims}
    \item[] Question: Do the main claims made in the abstract and introduction accurately reflect the paper's contributions and scope?
    \item[] Answer: \answerYes{} 
    \item[] Justification: The contributions and scope of the paper are summarized in the last paragraph of \cref{sec:introduction}.
    \item[] Guidelines:
    \begin{itemize}
      \item The answer NA means that the abstract and introduction do not include the claims made in the paper.
      \item The abstract and/or introduction should clearly state the claims made, including the contributions made in the paper and important assumptions and limitations. A No or NA answer to this question will not be perceived well by the reviewers. 
      \item The claims made should match theoretical and experimental results, and reflect how much the results can be expected to generalize to other settings. 
      \item It is fine to include aspirational goals as motivation as long as it is clear that these goals are not attained by the paper. 
    \end{itemize}
  \item {\bf Limitations}
    \item[] Question: Does the paper discuss the limitations of the work performed by the authors?
    \item[] Answer: \answerYes{} 
    \item[] Justification: The limitations of the work is discussed in \cref{app:limitations}.
    \item[] Guidelines:
    \begin{itemize}
      \item The answer NA means that the paper has no limitation while the answer No means that the paper has limitations, but those are not discussed in the paper. 
      \item The authors are encouraged to create a separate "Limitations" section in their paper.
      \item The paper should point out any strong assumptions and how robust the results are to violations of these assumptions (e.g., independence assumptions, noiseless settings, model well-specification, asymptotic approximations only holding locally). The authors should reflect on how these assumptions might be violated in practice and what the implications would be.
      \item The authors should reflect on the scope of the claims made, e.g., if the approach was only tested on a few datasets or with a few runs. In general, empirical results often depend on implicit assumptions, which should be articulated.
      \item The authors should reflect on the factors that influence the performance of the approach. For example, a facial recognition algorithm may perform poorly when image resolution is low or images are taken in low lighting. Or a speech-to-text system might not be used reliably to provide closed captions for online lectures because it fails to handle technical jargon.
      \item The authors should discuss the computational efficiency of the proposed algorithms and how they scale with dataset size.
      \item If applicable, the authors should discuss possible limitations of their approach to address problems of privacy and fairness.
      \item While the authors might fear that complete honesty about limitations might be used by reviewers as grounds for rejection, a worse outcome might be that reviewers discover limitations that aren't acknowledged in the paper. The authors should use their best judgment and recognize that individual actions in favor of transparency play an important role in developing norms that preserve the integrity of the community. Reviewers will be specifically instructed to not penalize honesty concerning limitations.
    \end{itemize}
  \item {\bf Theory Assumptions and Proofs}
    \item[] Question: For each theoretical result, does the paper provide the full set of assumptions and a complete (and correct) proof?
    \item[] Answer: \answerNA{} 
    \item[] Justification: The paper does not include theoretical results.
    \item[] Guidelines:
    \begin{itemize}
      \item The answer NA means that the paper does not include theoretical results. 
      \item All the theorems, formulas, and proofs in the paper should be numbered and cross-referenced.
      \item All assumptions should be clearly stated or referenced in the statement of any theorems.
      \item The proofs can either appear in the main paper or the supplemental material, but if they appear in the supplemental material, the authors are encouraged to provide a short proof sketch to provide intuition. 
      \item Inversely, any informal proof provided in the core of the paper should be complemented by formal proofs provided in appendix or supplemental material.
      \item Theorems and Lemmas that the proof relies upon should be properly referenced. 
    \end{itemize}
  \item {\bf Experimental Result Reproducibility}
    \item[] Question: Does the paper fully disclose all the information needed to reproduce the main experimental results of the paper to the extent that it affects the main claims and/or conclusions of the paper (regardless of whether the code and data are provided or not)?
    \item[] Answer: \answerYes{} 
    \item[] Justification: The model architecture is demonstrated in \cref{sec:two_stage_image_generation}. The implementation details are illustrated in \cref{app:implementation_details}. The code is released at \GitHub.
    \item[] Guidelines:
    \begin{itemize}
      \item The answer NA means that the paper does not include experiments.
      \item If the paper includes experiments, a No answer to this question will not be perceived well by the reviewers: Making the paper reproducible is important, regardless of whether the code and data are provided or not.
      \item If the contribution is a dataset and/or model, the authors should describe the steps taken to make their results reproducible or verifiable. 
      \item Depending on the contribution, reproducibility can be accomplished in various ways. For example, if the contribution is a novel architecture, describing the architecture fully might suffice, or if the contribution is a specific model and empirical evaluation, it may be necessary to either make it possible for others to replicate the model with the same dataset, or provide access to the model. In general. releasing code and data is often one good way to accomplish this, but reproducibility can also be provided via detailed instructions for how to replicate the results, access to a hosted model (e.g., in the case of a large language model), releasing of a model checkpoint, or other means that are appropriate to the research performed.
      \item While NeurIPS does not require releasing code, the conference does require all submissions to provide some reasonable avenue for reproducibility, which may depend on the nature of the contribution. For example
      \begin{enumerate}
        \item If the contribution is primarily a new algorithm, the paper should make it clear how to reproduce that algorithm.
        \item If the contribution is primarily a new model architecture, the paper should describe the architecture clearly and fully.
        \item If the contribution is a new model (e.g., a large language model), then there should either be a way to access this model for reproducing the results or a way to reproduce the model (e.g., with an open-source dataset or instructions for how to construct the dataset).
        \item We recognize that reproducibility may be tricky in some cases, in which case authors are welcome to describe the particular way they provide for reproducibility. In the case of closed-source models, it may be that access to the model is limited in some way (e.g., to registered users), but it should be possible for other researchers to have some path to reproducing or verifying the results.
      \end{enumerate}
    \end{itemize}
  \item {\bf Open access to data and code}
    \item[] Question: Does the paper provide open access to the data and code, with sufficient instructions to faithfully reproduce the main experimental results, as described in supplemental material?
    \item[] Answer: \answerYes{} 
    \item[] Justification: The code is released at \GitHub. The data used are publicly available.
    \item[] Guidelines:
    \begin{itemize}
      \item The answer NA means that paper does not include experiments requiring code.
      \item Please see the NeurIPS code and data submission guidelines (\url{https://nips.cc/public/guides/CodeSubmissionPolicy}) for more details.
      \item While we encourage the release of code and data, we understand that this might not be possible, so “No” is an acceptable answer. Papers cannot be rejected simply for not including code, unless this is central to the contribution (e.g., for a new open-source benchmark).
      \item The instructions should contain the exact command and environment needed to run to reproduce the results. See the NeurIPS code and data submission guidelines (\url{https://nips.cc/public/guides/CodeSubmissionPolicy}) for more details.
      \item The authors should provide instructions on data access and preparation, including how to access the raw data, preprocessed data, intermediate data, and generated data, etc.
      \item The authors should provide scripts to reproduce all experimental results for the new proposed method and baselines. If only a subset of experiments are reproducible, they should state which ones are omitted from the script and why.
      \item At submission time, to preserve anonymity, the authors should release anonymized versions (if applicable).
      \item Providing as much information as possible in supplemental material (appended to the paper) is recommended, but including URLs to data and code is permitted.
    \end{itemize}
  \item {\bf Experimental Setting/Details}
    \item[] Question: Does the paper specify all the training and test details (e.g., data splits, hyperparameters, how they were chosen, type of optimizer, etc.) necessary to understand the results?
    \item[] Answer: \answerYes{} 
    \item[] Justification: Implementation details are illustrated in \cref{app:implementation_details}.
    \item[] Guidelines:
    \begin{itemize}
      \item The answer NA means that the paper does not include experiments.
      \item The experimental setting should be presented in the core of the paper to a level of detail that is necessary to appreciate the results and make sense of them.
      \item The full details can be provided either with the code, in appendix, or as supplemental material.
    \end{itemize}
  \item {\bf Experiment Statistical Significance}
    \item[] Question: Does the paper report error bars suitably and correctly defined or other appropriate information about the statistical significance of the experiments?
    \item[] Answer: \answerNo{} 
    \item[] Justification: Given the massive amount of experiments conducted in this paper, providing error bars would be computationally prohibitive.
    \item[] Guidelines:
    \begin{itemize}
      \item The answer NA means that the paper does not include experiments.
      \item The authors should answer "Yes" if the results are accompanied by error bars, confidence intervals, or statistical significance tests, at least for the experiments that support the main claims of the paper.
      \item The factors of variability that the error bars are capturing should be clearly stated (for example, train/test split, initialization, random drawing of some parameter, or overall run with given experimental conditions).
      \item The method for calculating the error bars should be explained (closed form formula, call to a library function, bootstrap, etc.)
      \item The assumptions made should be given (e.g., Normally distributed errors).
      \item It should be clear whether the error bar is the standard deviation or the standard error of the mean.
      \item It is OK to report 1-sigma error bars, but one should state it. The authors should preferably report a 2-sigma error bar than state that they have a 96\% CI, if the hypothesis of Normality of errors is not verified.
      \item For asymmetric distributions, the authors should be careful not to show in tables or figures symmetric error bars that would yield results that are out of range (e.g. negative error rates).
      \item If error bars are reported in tables or plots, The authors should explain in the text how they were calculated and reference the corresponding figures or tables in the text.
    \end{itemize}
  \item {\bf Experiments Compute Resources}
    \item[] Question: For each experiment, does the paper provide sufficient information on the computer resources (type of compute workers, memory, time of execution) needed to reproduce the experiments?
    \item[] Answer: \answerYes{} 
    \item[] Justification: Compute resources are described in \cref{app:implementation_details}.
    \item[] Guidelines:
    \begin{itemize}
      \item The answer NA means that the paper does not include experiments.
      \item The paper should indicate the type of compute workers CPU or GPU, internal cluster, or cloud provider, including relevant memory and storage.
      \item The paper should provide the amount of compute required for each of the individual experimental runs as well as estimate the total compute. 
      \item The paper should disclose whether the full research project required more compute than the experiments reported in the paper (e.g., preliminary or failed experiments that didn't make it into the paper). 
    \end{itemize}
  \item {\bf Code Of Ethics}
    \item[] Question: Does the research conducted in the paper conform, in every respect, with the NeurIPS Code of Ethics \url{https://neurips.cc/public/EthicsGuidelines}?
    \item[] Answer: \answerYes{} 
    \item[] Justification: The paper conforms with the NeurIPS Code of Ethics.
    \item[] Guidelines:
    \begin{itemize}
      \item The answer NA means that the authors have not reviewed the NeurIPS Code of Ethics.
      \item If the authors answer No, they should explain the special circumstances that require a deviation from the Code of Ethics.
      \item The authors should make sure to preserve anonymity (e.g., if there is a special consideration due to laws or regulations in their jurisdiction).
    \end{itemize}
  \item {\bf Broader Impacts}
    \item[] Question: Does the paper discuss both potential positive societal impacts and negative societal impacts of the work performed?
    \item[] Answer: \answerYes{} 
    \item[] Justification: Potential societal impacts of the work are discussed in \cref{app:broader_impacts}.
    \item[] Guidelines:
    \begin{itemize}
      \item The answer NA means that there is no societal impact of the work performed.
      \item If the authors answer NA or No, they should explain why their work has no societal impact or why the paper does not address societal impact.
      \item Examples of negative societal impacts include potential malicious or unintended uses (e.g., disinformation, generating fake profiles, surveillance), fairness considerations (e.g., deployment of technologies that could make decisions that unfairly impact specific groups), privacy considerations, and security considerations.
      \item The conference expects that many papers will be foundational research and not tied to particular applications, let alone deployments. However, if there is a direct path to any negative applications, the authors should point it out. For example, it is legitimate to point out that an improvement in the quality of generative models could be used to generate deepfakes for disinformation. On the other hand, it is not needed to point out that a generic algorithm for optimizing neural networks could enable people to train models that generate Deepfakes faster.
      \item The authors should consider possible harms that could arise when the technology is being used as intended and functioning correctly, harms that could arise when the technology is being used as intended but gives incorrect results, and harms following from (intentional or unintentional) misuse of the technology.
      \item If there are negative societal impacts, the authors could also discuss possible mitigation strategies (e.g., gated release of models, providing defenses in addition to attacks, mechanisms for monitoring misuse, mechanisms to monitor how a system learns from feedback over time, improving the efficiency and accessibility of ML).
    \end{itemize}
  \item {\bf Safeguards}
    \item[] Question: Does the paper describe safeguards that have been put in place for responsible release of data or models that have a high risk for misuse (e.g., pretrained language models, image generators, or scraped datasets)?
    \item[] Answer: \answerNA{} 
    \item[] Justification: The paper poses no such risks.
    \item[] Guidelines:
    \begin{itemize}
      \item The answer NA means that the paper poses no such risks.
      \item Released models that have a high risk for misuse or dual-use should be released with necessary safeguards to allow for controlled use of the model, for example by requiring that users adhere to usage guidelines or restrictions to access the model or implementing safety filters. 
      \item Datasets that have been scraped from the Internet could pose safety risks. The authors should describe how they avoided releasing unsafe images.
      \item We recognize that providing effective safeguards is challenging, and many papers do not require this, but we encourage authors to take this into account and make a best faith effort.
    \end{itemize}
  \item {\bf Licenses for existing assets}
    \item[] Question: Are the creators or original owners of assets (e.g., code, data, models), used in the paper, properly credited and are the license and terms of use explicitly mentioned and properly respected?
    \item[] Answer: \answerYes{} 
    \item[] Justification: The paper mainly uses the \BEiTvTwo{} repository, which is cited in the paper and mentioned in the code.
    \item[] Guidelines:
    \begin{itemize}
      \item The answer NA means that the paper does not use existing assets.
      \item The authors should cite the original paper that produced the code package or dataset.
      \item The authors should state which version of the asset is used and, if possible, include a URL.
      \item The name of the license (e.g., CC-BY 4.0) should be included for each asset.
      \item For scraped data from a particular source (e.g., website), the copyright and terms of service of that source should be provided.
      \item If assets are released, the license, copyright information, and terms of use in the package should be provided. For popular datasets, \url{paperswithcode.com/datasets} has curated licenses for some datasets. Their licensing guide can help determine the license of a dataset.
      \item For existing datasets that are re-packaged, both the original license and the license of the derived asset (if it has changed) should be provided.
      \item If this information is not available online, the authors are encouraged to reach out to the asset's creators.
    \end{itemize}
  \item {\bf New Assets}
    \item[] Question: Are new assets introduced in the paper well documented and is the documentation provided alongside the assets?
    \item[] Answer: \answerNA{} 
    \item[] Justification: The paper does not release new assets.
    \item[] Guidelines:
    \begin{itemize}
      \item The answer NA means that the paper does not release new assets.
      \item Researchers should communicate the details of the dataset/code/model as part of their submissions via structured templates. This includes details about training, license, limitations, etc. 
      \item The paper should discuss whether and how consent was obtained from people whose asset is used.
      \item At submission time, remember to anonymize your assets (if applicable). You can either create an anonymized URL or include an anonymized zip file.
    \end{itemize}
  \item{\bf Crowdsourcing and Research with Human Subjects}
    \item[] Question: For crowdsourcing experiments and research with human subjects, does the paper include the full text of instructions given to participants and screenshots, if applicable, as well as details about compensation (if any)? 
    \item[] Answer: \answerNA{} 
    \item[] Justification: The paper does not involve crowdsourcing nor research with human subjects.
    \item[] Guidelines:
    \begin{itemize}
      \item The answer NA means that the paper does not involve crowdsourcing nor research with human subjects.
      \item Including this information in the supplemental material is fine, but if the main contribution of the paper involves human subjects, then as much detail as possible should be included in the main paper. 
      \item According to the NeurIPS Code of Ethics, workers involved in data collection, curation, or other labor should be paid at least the minimum wage in the country of the data collector. 
    \end{itemize}
  \item{\bf Institutional Review Board (IRB) Approvals or Equivalent for Research with Human Subjects}
    \item[] Question: Does the paper describe potential risks incurred by study participants, whether such risks were disclosed to the subjects, and whether Institutional Review Board (IRB) approvals (or an equivalent approval/review based on the requirements of your country or institution) were obtained?
    \item[] Answer: \answerNA{} 
    \item[] Justification: The paper does not involve crowdsourcing nor research with human subjects.
    \item[] Guidelines:
    \begin{itemize}
      \item The answer NA means that the paper does not involve crowdsourcing nor research with human subjects.
      \item Depending on the country in which research is conducted, IRB approval (or equivalent) may be required for any human subjects research. If you obtained IRB approval, you should clearly state this in the paper. 
      \item We recognize that the procedures for this may vary significantly between institutions and locations, and we expect authors to adhere to the NeurIPS Code of Ethics and the guidelines for their institution. 
      \item For initial submissions, do not include any information that would break anonymity (if applicable), such as the institution conducting the review.
    \end{itemize}

\end{enumerate}

\end{document}